\documentclass[10pt,twocolumn,letterpaper]{article}

\usepackage[pagenumbers]{cvpr} 
\usepackage{makecell}  
\usepackage{multirow}
\usepackage[table,xcdraw]{xcolor}
\definecolor{cvprblue}{rgb}{0.21,0.49,0.74}
\usepackage[pagebackref,breaklinks,colorlinks,allcolors=cvprblue]{hyperref}

\def\paperID{****} 
\def\confName{CVPR}
\def\confYear{2026}

\title{Visual Instruction Pretraining for Domain-Specific Foundation Models}

\author{Yuxuan Li, Yicheng Zhang, Wenhao Tang, Ming-Ming Cheng, Yimian Dai, \setcounter{footnote}{1}
Xiang Li\thanks{Corresponding author: Xiang Li and Jian Yang. \\
All authors are with PCA Lab, VCIP, Computer Science, NKU,
Tianjin, China. E-mails: yuxuan.li.17@alumni.ucl.ac.uk; \{whtang, zhangyc\}@mail.nankai.edu.cn; \{yimian.dai, cmm, xiang.li.implus, csjyang\}@nankai.edu.cn. \\
Ming-Ming Cheng and Xiang Li are also with NKIARI, Futian, Shenzhen, China. \\
This work has been submitted to the IEEE for possible publication. Copyright may be transferred without notice, after which this version may no longer be accessible.}, Jian Yang\footnotemark[2]  \\
}

\begin{document}
\maketitle
\begin{abstract}
Human visual perception is the product of a synergistic interplay between bottom-up perception processing and top-down understanding-driven modulation. While modern computer vision has excelled at creating powerful models that mirror the bottom-up, hierarchical nature of perception, the reciprocal top-down influence of high-level understanding on the foundational learning of low-level perception features remains critically underexplored. This paper addresses this gap by proposing a new paradigm for pretraining foundation models, inspired by the top-down mechanisms of human vision. We introduce \textbf{V}isual \textbf{i}ns\textbf{T}ruction \textbf{P}retraining (ViTP), a novel approach that directly leverages understanding to enhance perception. ViTP embeds a Vision Transformer backbone within a Vision-Language Model and pretrains it end-to-end using a rich corpus of visual instruction data curated from target downstream domains. ViTP is powered by our proposed Visual Robustness Learning (VRL), which compels the Vision Transformer to learn robust and domain-relevant features from a sparse set of visual tokens. Extensive experiments on 16 challenging remote sensing and medical imaging benchmarks demonstrate that ViTP establishes new state-of-the-art performance across a diverse range of downstream tasks. The code is available at \url{https://github.com/zcablii/ViTP}
\end{abstract}
 
\section{Introduction}

Decades of research in cognitive neuroscience have established that human visual perception is not a one-way street. It is a dynamic process arising from the constant interaction of two fundamental pathways: a bottom-up, data-driven pathway and a top-down, knowledge-driven one. The bottom-up process, famously elucidated by Hubel and Wiesel~\cite{hubel1962}, involves a hierarchical construction of understanding: simple features like edges and orientations are extracted by early visual cortices and progressively assembled into complex object representations in higher cortical areas~\cite{marr2010vision, biederman1987}. This paradigm has been the primary inspiration for modern computer vision. For instance, prevailing Vision Transformer pretraining paradigms include supervised classification, MIM \cite{mae,beit,scalemae}, contrastive learning \cite{moco,dinov2,skysense}, and image-text contrastive learning \cite{clip,remoteclip,rs5m}. 
Recent multimodal large models~\cite{qwen,internvl} usually utilize a vision tower to provide vision elements for high-level understanding. These paradigms are all fundamentally ``bottom-up'': they operate on the premise that robust low-level perception is a prerequisite for high-level understanding \cite{Swin,lin2017feature}. 

\begin{figure}[!t]
\centering
\includegraphics[width=0.9\linewidth]{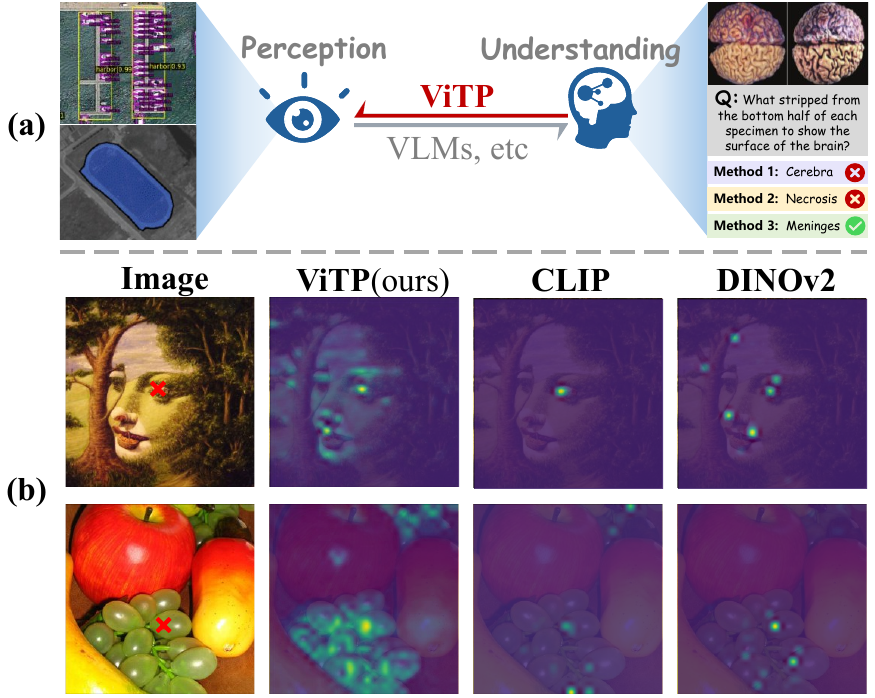}
\vspace{-8pt}
\caption{(a) The synergistic relationship between perception and understanding in modern CV. Our proposed ViTP forges a previously underexplored link from high-level understanding to low-level perception. (b) Self-attention activation maps for the query patch (marked with a red cross). ViTP identifies fine-grained object parts that are high-level semantically related. }
\label{fig:1}
\vspace{-14pt}
\end{figure}
 
However, this bottom-up hierarchy only tells half the story. A wealth of research in cognitive neuroscience has demonstrated the profound top-down influence of high-level understanding, attention, and prior knowledge on foundational perception. Cognitive processes, guided by expectation and goals, actively modulate and refine how low-level visual information is processed, even at the earliest stages in the visual cortex~\cite{kastner1998, rao1999predictive, muckli2015contextual}. This feedback mechanism allows the brain to focus on relevant details, fill in missing information, and interpret ambiguous stimuli with remarkable efficiency. While this reciprocal pathway—where understanding enhances perception—is integral to human vision, it remains critically underexplored in the computer vision models. This raises a compelling question: can abstract, high-level understanding be harnessed to directly guide a perception model toward learning more precise and domain-relevant representations?

\begin{figure}[!t]
\centering
\includegraphics[width=0.91\linewidth]{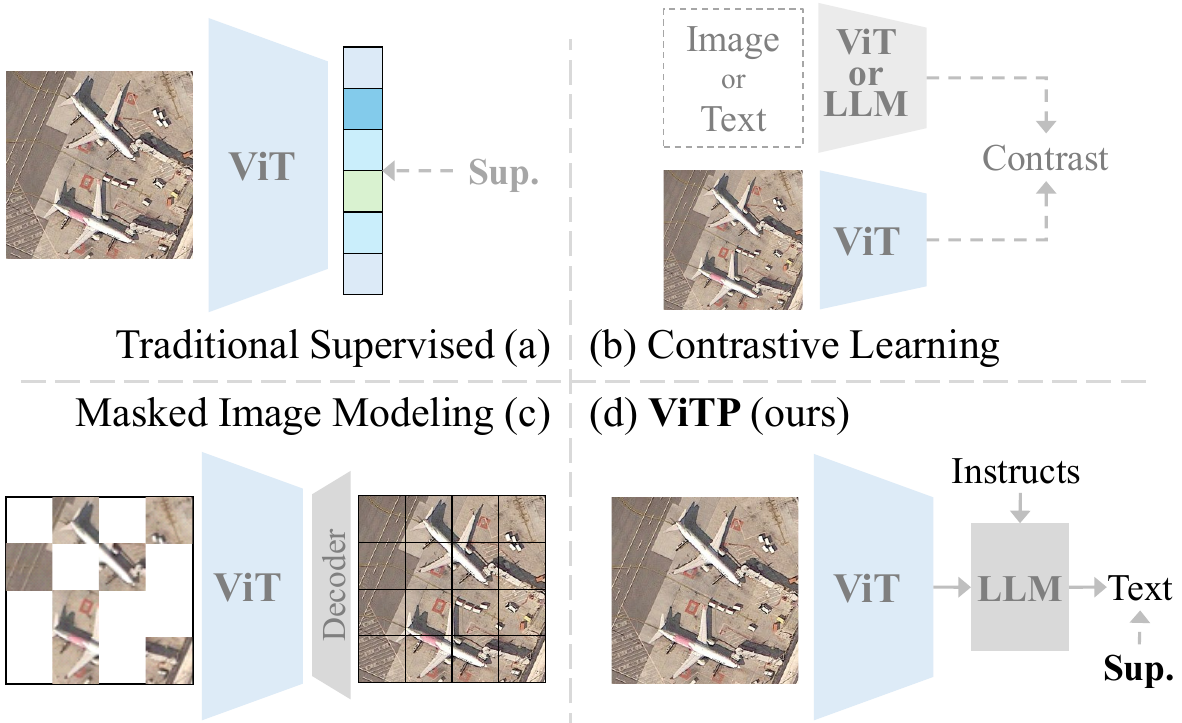}
\vspace{-10pt}
\caption{Comparison of pretraining paradigms for Vision Transformer (ViT) foundation models. ViTP employs an instruction-following objective to directly instil domain-specific perception capabilities into the vision backbone.}
\vspace{-12pt}
\label{fig:task_comp}
\end{figure}

Answering this question is particularly critical for pretraining powerful foundation models in specialized domains like remote sensing and medical imaging, where prevailing paradigms face formidable challenges. 
The supervised~\cite{deit3} pretraining risks overfitting to narrow label distributions; the MIM~\cite{mae} may inadvertently neglect the fine-grained details of small but critical objects; image-image contrastive methods, while effective, are often notoriously difficult to optimize and demand substantial computational resources. 
Prevailing pretraining paradigms of image-to-text contrastive learning~\cite{clip} or multimodal autoregressive Vision Transformer pretraining~\cite{aim} represent steps toward reasoning-guided pretraining. 
However, these methods primarily aim for general-purpose image-text alignment rather than optimizing a vision backbone for specialized, fine-grained downstream tasks. Consequently, the global, image-level features learned by models like CLIP~\cite{clip} or AIM~\cite{aim} often prove suboptimal for dense prediction tasks such as semantic segmentation or object detection~\cite{densevlm}. This limitation underscores that merely aligning global image and text representations is an insufficient strategy for forging a powerful, domain-specific perception backbone.

To bridge this gap, we introduce \textbf{V}isual \textbf{i}ns\textbf{T}ruction \textbf{P}retraining (ViTP), a novel, top-down pretraining paradigm that directly integrates high-level, instruction-based understanding into the perceptual feature learning process of a Vision Transformer backbone. As illustrated in Figure~\ref{fig:task_comp}, ViTP diverges from prior methods by embedding the Vision Transformer within a larger Vision-Language Model (VLM) and pretraining it via a visual instruction-following objective. We construct our training data with image-text pairs derived directly from the target downstream tasks. During training, image tokens from the Vision Transformer and text tokens from a user query are fed jointly into a Large Language Model (LLM) \cite{llava}. The LLM's generated response serves as the supervisory signal, compelling the Vision Transformer backbone to learn the complex data distributions of the downstream domain in an end-to-end fashion. This process endows the Vision Transformer with highly relevant, domain-specific perceptual capabilities. Furthermore, we introduce \textbf{V}isual \textbf{R}obustness \textbf{L}earning (VRL) by randomly dropping a large fraction of the Vision Transformer's output image tokens before they are passed to the LLM. This constraint implicitly forces the Vision Transformer's attention mechanism to encode more comprehensive and robust information within each of the remaining tokens, thereby enhancing the robustness and semantic richness of the learned visual features.

\begin{figure}[!t]
\centering
\includegraphics[width=0.9\linewidth]{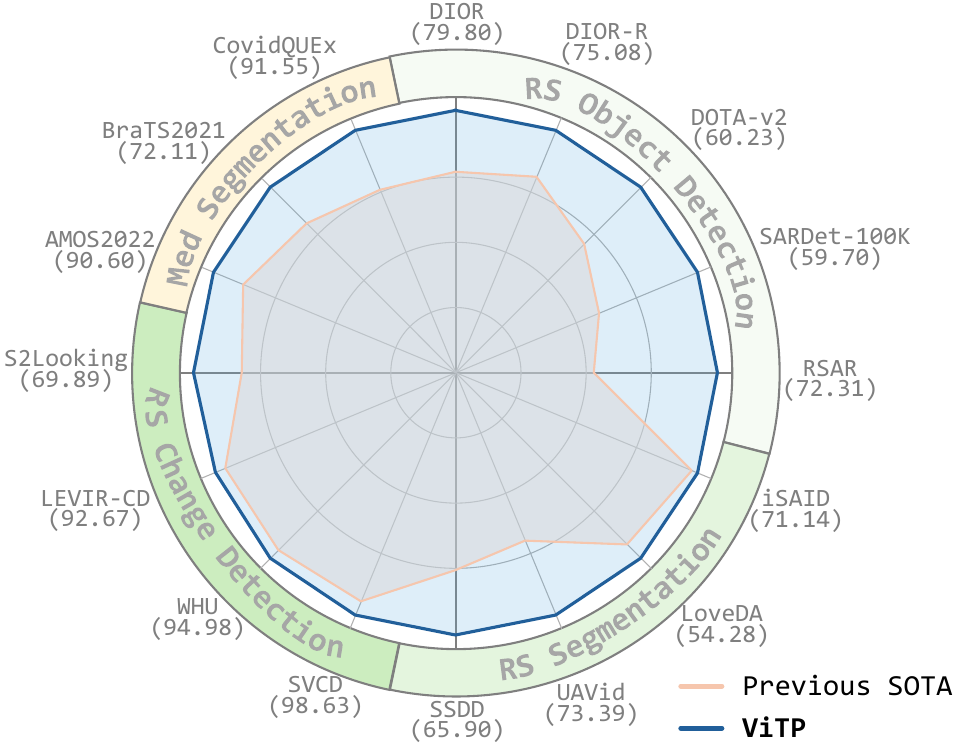}
\vspace{-6pt}
\caption{ViTP sets new SOTA performance across a diverse range of downstream tasks in medical imaging and remote sensing.}
\vspace{-12pt}
\label{fig:radar}
\end{figure}

Extensive experiments on 16 challenging remote sensing and medical imaging benchmarks validate the effectiveness and efficiency of our approach. As shown in Figure~\ref{fig:radar}, ViTP achieves new state-of-the-art results across several tasks. Notably, the ViTP pretraining process is computationally efficient, requiring only one day on 8 A40 GPUs. This work not only presents a novel pretraining paradigm but also offers a promising solution for creating powerful, domain-adapted foundation models.
 
Our contributions are summarized as follows: 
\begin{itemize}
    \item We introduce ViTP, the first top-down pretraining paradigm that leverages understanding tasks to imbue a Vision Transformer backbone with high-level semantic perception 
    , as illustrated in Figure~\ref{fig:1}.
    \item We propose Visual Robustness Learning, a regularization method that encourages the Vision Transformer to learn more comprehensive and robust feature representations by operating on a sparse set of visual tokens.
    \item Extensive experiments demonstrate the pretraining efficiency and state-of-the-art performance of ViTP on downstream tasks in remote sensing and medical imaging.
\end{itemize}

\section{Related Work}
\label{sec:related}

\subsection{Perception Foundation Model}

Early approaches predominantly relied on supervised pretraining over large-scale labeled datasets such as ImageNet~\cite{resnet,vit}. While effective in learning semantically meaningful representations, these methods risk overfitting to the label space of the pretraining task, thereby limiting generalization to diverse downstream applications. Consequently, the field has increasingly shifted toward unsupervised learning paradigms that leverage unlabeled data. Among these, contrastive learning and masked image modeling (MIM) emerged as two dominant strategies. 
Contrastive learning methods, including MoCo~\cite{moco}, BYOL~\cite{byol}, and DINOv2~\cite{dinov2}, learn discriminative representations by minimizing the distance between positive pairs (e.g., augmented views of the same image) while maximizing the distance between negative pairs in the embedding space. These methods emphasize semantic invariance and shown strong generalization capabilities across tasks. On the other hand, MIM-based approaches such as MAE~\cite{mae}, SimMIM~\cite{simmim} and UM-MAE~\cite{ummae} adopt a generative pretext task: a large portion of the input image is masked, and the model is trained to reconstruct the missing content. This encourages the model to learn holistic, contextually rich visual features by capturing spatial dependencies and fine-grained details.
While both paradigms achieved impressive results on natural images, their direct transfer to specialized domains such as remote sensing or medical imagery often results in suboptimal performance~\cite{satmae,remoteclip,medmae}. This underscores the necessity for domain-specific adaptations in pretraining strategies.

\subsection{Visual Instruction Tuning}
The rise of powerful VLMs such as Gemini~\cite{gemini}, InternVL~\cite{internvl}, and Qwen-VL~\cite{qwen25vl} are largely driven by visual instruction tuning~\cite{llava}. This paradigm aims to align visual representations with a pretrained Large Language Model by projecting image features into the LLM’s embedding space. The model is then finetuned on a corpus of image-text instruction pairs, enabling the LLM to interpret visual inputs and respond to textual instructions. 
A key challenge in this paradigm is the scarcity of high-quality, large-scale visual instruction data. Foundational models like LLaVA~\cite{llava} addressed this by creating synthetic datasets: they used a powerful language model (e.g., GPT-4~\cite{gpt4}) to generate diverse and complex reasoning-based datasets.
In the visual instruction tuning standard setup, the vision encoder typically remains frozen, serving solely as a feature extractor, while all learning is concentrated in the LLM and the projection module.
While this paradigm effectively leverages perception to enhance understanding, the reverse direction, i.e. using understanding to improve perception, remains largely underexplored. Our proposed ViTP addresses this gap by inverting the traditional visual instruction tuning pipeline. Instead of using images to tune an LLM, we utilize an LLM to guide the pretraining of the vision encoder.

More related work is provided in the \textbf{Supplementary Material}.
\section{Method}
\begin{figure*}[t]
  \centering
  \includegraphics[width=0.9\linewidth]{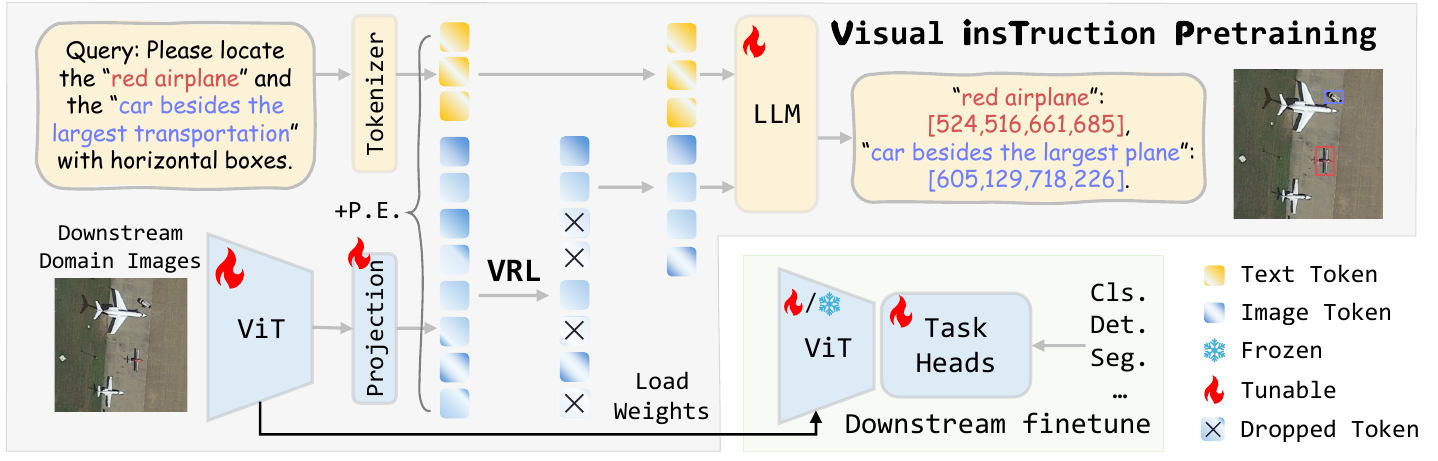}
  \caption{A conceptual illustration of the ViTP framework. A Vision Transformer (ViT) backbone is embedded within a large VLM and then pretrained with domain-specific instruction following objective and Visual Robustness Learning (VRL). This process instils high-level semantic understanding into the ViT. The resulting weights are then used to initialize models for various downstream perception tasks.}
  \label{fig:ViTP}
\end{figure*}

In this section, we present our visual instruction pretraining (ViTP) framework, a novel paradigm designed to pretrain a ViT backbone. As illustrated in Figure~\ref{fig:task_comp}, ViTP fundamentally differs from traditional pretraining by leveraging the understanding capabilities of modern VLMs. The entire process is driven by a ``visual instruction following objective'', where the ViT learns to extract features that help an LLM answer questions about an image. To tailor the model for specific downstream applications, we outline a ``data recipe'' for curating a domain-specific pretraining dataset. During pretraining, we employ ``Visual Robustness Learning (VRL)'', a regularization technique that drops image tokens to enhance the semantic robustness of the learned representations. The final pretrained ViT serves as a powerful foundation model, readily adaptable to a variety of downstream tasks.

\subsection{Visual Instruction Following Objective}
\label{sec:pretrain_paradigm}
The central hypothesis of ViTP is that a ViT can learn relevant and expressive features if its training is guided by high-level understanding. To achieve this, we frame the pretraining as a visual instruction-following task. The framework, depicted in Figure~\ref{fig:ViTP}, processes a domain-specific image through a ViT encoder to produce a sequence of image tokens. These tokens are projected into the LLM's embedding space and concatenated with the tokenized text of an instruction. A Large Language Model (LLM) then processes this combined sequence to generate a response. The entire model is trained end-to-end, allowing the supervisory signal from the LLM's response to directly optimize the ViT's feature extraction process.

Our approach employs a continual pretraining strategy~\cite{dontstop}, starting with a well-trained, general-purpose VLM. We then continue its training on our curated domain-specific datasets. Such a continual pretraining strategy leverages the VLM's vast pre-existing knowledge of general visual and linguistic patterns, providing a robust initialization. It significantly accelerates convergence during domain-specific pretraining, enhancing computational efficiency. Let the raw dataset be $\mathcal{D}_{\text{raw}} = \{(I, Q, R)\}$, where $I$ is an image, $Q$ is a text query (instruction), and $R$ is the ground-truth text response. This is processed into the final training set $\mathcal{D} = \{(x_i, x_t, y^\star)\}$, where $x_i$, $x_t$, and $y^\star$ represent the processed image tokens, text tokens, and target response tokens, respectively.

\subsubsection{ViT Feature Extraction and Projection}

Given an input image $I \in \mathbb{R}^{H \times W \times 3}$, the ViT backbone partitions it into a grid of non-overlapping patches. Each patch is linearly embedded, and these patch embeddings are prepended with a `[CLS]' token and processed through a series of Transformer blocks \cite{transformer}. This yields a sequence of output image tokens $x'_{i} = \{t_1, t_2, \ldots, t_N\}$, where $N$ is the sequence length. To align these visual tokens with the LLM's embedding space, a lightweight projection layer (e.g., a two-layer MLP) maps these visual tokens $x'_{i}$ into the final image tokens $x_{i}$.

\subsubsection{Instruction-following Token Concatenation}
For each image, the corresponding text query $Q$ is converted into a sequence of text tokens $x_{t}$ using the LLM's tokenizer. These tokens represent the task-specific instruction. The projected image tokens $x_{i}$ and text tokens $x_{t}$ are then concatenated to form a unified input sequence. Crucially, learnable positional encodings are added to the embeddings of the image and text tokens to provide the LLM with spatial and sequential context. The final input sequence for the LLM is formed as:
\begin{equation}
S_{llm} = [ \text{PE}(x_{i}) \,;\, \text{PE}(x_{t}) ]
\end{equation}
where $\text{PE}(\cdot)$ denotes the addition of positional encodings to the token embeddings and `$[;]$' signifies sequence concatenation.

\subsubsection{LLM-based Supervision}
The combined sequence $S_{llm}$ is processed by the LLM, which acts as a understanding engine to interpret the visual information from $x_i$ in the context of the instruction $x_t$. The model then auto-regressively generates an output sequence $O$. The entire system is optimized by minimizing the discrepancy between the generated output $O$ and the ground-truth target response $y^\star$. For instance, if the instruction asks to identify an object, $y^\star$ could be a textual description including its location in a structured format. During pretraining, the gradients from the output loss propagate back through the entire model, including the projection layer and the ViT backbone. We allow the weights of the ViT, the projection layer, and the LLM to be trainable. The optimization follows a standard supervised finetuning (SFT) objective, which minimizes the negative log-likelihood of the target sequence:
\begin{equation}
\mathcal{L}_{\mathrm{SFT}}(\theta)
=
\mathbb{E}
\Bigl[
-\log P_\theta \bigl( y^\star \,\big|\, S_{llm} \bigr)
\Bigr],
\label{eq:sft-sentence-loss}
\end{equation}
where \((x_i, x_t, y^\star) \sim \mathcal{D}\) and $P_\theta$ is the probability distribution over the text sequences parameterized by the entire model $\theta$.

\subsection{Visual Robustness Learning}
\label{sec:vrl}
To foster the learning of more robust and semantically rich features, we introduce Visual Robustness Learning (VRL), a simple yet effective regularization technique applied during pretraining. As shown in Figure~\ref{fig:ViTP}, VRL randomly drops a significant fraction of the projected image tokens $x_i$ before they are concatenated with the text tokens $x_t$. This operation is performed after positional encodings are associated with the tokens, ensuring the LLM retains knowledge of the original spatial positions of the surviving tokens. The VRL objective is thus:
\begin{equation}
\mathcal{L}_{\mathrm{VRL}}(\theta)
=
\mathbb{E}
\Bigl[
-\log P_\theta \bigl( y^\star \,\big|\,[ \mathcal{C}_r(\text{PE}(x_i)); \text{PE}(x_t) ]\bigr)
\Bigr],
\label{eq:vrl-sentence-loss}
\end{equation}
where $\mathcal{C}_r$ is a random sampling operation that drops a proportion $r$ of the tokens from a sequence. Formally, for a sequence $\mathcal{S}$:
\begin{equation}
\mathcal{C}_r(\mathcal{S}) \sim \bigl\{ X \subseteq \mathcal{S} \;\big|\; |X| = \lceil (1-r) \cdot |\mathcal{S}| \rceil \bigr\},
\end{equation}
where the subset $X$ is selected uniformly at random while preserving the original ordering, and $r \in [0,1)$ . This ``torture'' mechanism forces the ViT to encode more comprehensive information in each token, as the model must infer the full visual context from a partial input. It encourages the ViT's attention mechanism to learn robust, distributed and less redundant representations. As a practical benefit, dropping a large portion of tokens (e.g., $r=0.75$) significantly reduces memory usage and accelerates computation, enhancing ViTP's scalability.

\subsection{Pretraining Dataset Recipe}
\label{sec:dataset_recipe}
With the rapid development of VLMs, numerous image-text paired instruction datasets tailored for downstream domains have been released. The efficacy of ViTP is heavily dependent on the quality and composition of the pretraining dataset. We establish four key principles for constructing our data mixture:
\begin{enumerate}
  \item \textbf{Scale and Diversity:} The dataset must be large and diverse, containing a wide array of visual concepts, scenes, and objects representative of the target domain.
  \item \textbf{Modality Coverage:} The data must encompass all imaging modalities expected in downstream tasks. For instance, if a downstream task uses remote sensing Synthetic Aperture Radar (SAR) imagery, the pretraining mix should include such modality data to ensure the model learns modality-specific features.
  \item \textbf{Task Capability Alignment:} The instruction-following tasks in the pretraining data should foster abilities required downstream. For instance, for downstream object detection tasks, including visual grounding and fine-grained VQA during pretraining empowers the backbone with localization and spatial understanding capabilities.
  \item \textbf{Preservation of Generality:} A certain fraction of general-domain natural images (e.g., from public VLM datasets) should be included. Domain-specific data can be limited in diversity, adding general data mitigates overfitting and prevents the model from losing its foundational ability to understand broad visual patterns.
\end{enumerate}

\subsection{Downstream Finetuning}
Once ViTP is complete, the pretrained ViT backbone is then extracted to serve as a powerful backbone network. For downstream applications, the pretrained ViT is integrated into a standard ViT-Adapter~\cite{vitadapter} and combined with task-specific heads. The entire model is then finetuned on the target downstream dataset. This transfer learning approach leverages the rich, instruction-aware representations learned during ViTP, enabling faster adaptation and superior performance on specialized tasks.

\section{Experiments}

This section offers a rigorous assessment of ViTP’s finetuning capabilities on various downstream tasks from both remote-sensing and medical imaging domains. 
We present our experimental setup, a thorough analysis of the results, and an ablation study. We demonstrate ViTP's superior performance compared to existing state-of-the-art methods and analyze its efficiency and robustness. The best score is indicated in \textbf{bold}, while the second-best score is \underline{underlined}. 

\textit{Implementation details, dataset descriptions, additional experiments on general-domain datasets, and extended analyzes are provided in the \textbf{Supplementary Material}.}

\subsection{Remote Sensing}

\begin{table}[!t]
\caption{Object detection performance (mAP \%) on optical remote sensing datasets.}
\label{tab:det_rgb_results}
\renewcommand\arraystretch{1.1}  
\setlength{\tabcolsep}{3.5pt}
  \centering \footnotesize
\begin{tabular}{c|cc||c|c}
\textbf{Model} & \textbf{DIOR} & \textbf{DIOR-R} & \textbf{Model} & \textbf{DOTA-v2} \\ \Xhline{1pt}
GASSL~\cite{gassl} & 67.40 & 65.65  & RetinaNet~\cite{retina} & 46.68 \\
SatMAE~\cite{satmae} & 70.89 &  62.30  & F-RCNN~\cite{fasterrcnn} & 47.31 \\
RingMo~\cite{RinMo} & 75.90 & - &  FCOS~\cite{fcos} & 48.51 \\
CACO~\cite{caco} & 66.91 & 64.10 & ATSS~\cite{atss} & 49.57 \\
SSL4EO~\cite{ssl4eo} & 64.82 & 61.23 & SASM~\cite{sasm} & 44.53 \\
CMID~\cite{cmid} & 75.11 & 66.37 & S2ANet~\cite{s2anet} & 49.86 \\
RVSA~\cite{rvsa} & 73.22 & 70.96 &  KLD~\cite{kld} & 47.26 \\
SatLas~\cite{satlas} & 74.10 & 67.59 & O-RepPoints~\cite{RepPoints} & 48.95 \\
GFM~\cite{gfm} & 72.84 & 67.67 &  RoT Trans.~\cite{roitrans} & 52.81 \\
ScaleMAE~\cite{scalemae} & 73.81 & 70.20 & O-RCNN~\cite{orcnn} & 53.28 \\
MA3E~\cite{ma3e} & - & 71.82 & GGHF~\cite{gghf} & 57.17 \\
Sel-MAE~\cite{selectivemae} & 78.70 & 71.75 &  DCFL~\cite{dcfl} & 57.66 \\ 
SkySense~\cite{skysense} & \underline{78.73} & \underline{74.27} & BillionFM~\cite{billionfm} & \underline{58.69} \\ \hline
\textbf{ViTP} & \textbf{79.80} & \textbf{75.08 }& \textbf{ViTP} & \textbf{60.23}
\end{tabular}
\end{table}

\subsubsection{\textbf{Object detection}}

\begin{table}[!t]
\caption{Object detection performance (mAP \%) on SAR datasets.}
\label{tab:det_sar_results}
\renewcommand\arraystretch{1.1}  
\setlength{\tabcolsep}{3pt}
  \centering \footnotesize
\begin{tabular}{cc|cc}
\multicolumn{1}{c}{\textbf{Model}} & \multicolumn{1}{c|}{\textbf{SARDet-100K}} & \multicolumn{1}{c}{\textbf{Model}} & \multicolumn{1}{c}{\textbf{RSAR}} \\ \Xhline{1pt}

DETR~\cite{detr} & 31.8 & Def. DETR~\cite{deformabledetr} & 46.62 \\
Sparse RCNN~\cite{sparsercnn} & 38.1 &   RetinaNet~\cite{retina}  & 57.67 \\
Dab-DETR~\cite{dabdetr} & 45.9 &  ARS-DETR~\cite{zeng2024ars} & 61.14\\
FCOS~\cite{fcos} & 46.5 &  R3Det~\cite{r3det} & 63.94\\
Grid RCNN~\cite{gridrcnn} & 48.8 & LLMRotate~\cite{llmrotate} & 64.1 \\
GFL~\cite{gfl} & 49.8 & ReDet~\cite{redet} & 64.71  \\
Defor. DETR~\cite{deformabledetr} & 50.0 & O-RCNN~\cite{orcnn} & 64.82\\
MSFA~\cite{sardet} & 53.7 & S2ANet~\cite{s2anet} & 66.47\\
DenoDet~\cite{denodet} & 55.4 & RoI-Trans.~\cite{roitrans} & 66.95 \\
DenoDetv2~\cite{denodetv2} & 56.4 & SatMAE~\cite{satmae} & 67.99  \\
SARATR-X~\cite{saratr} & \underline{57.3} &  RemoteCLIP~\cite{remoteclip} & \underline{69.18}\\ \hline
\textbf{ViTP} & \textbf{59.7} & \textbf{ViTP} & \textbf{72.31}
\end{tabular}
\end{table}

Table~\ref{tab:det_rgb_results} presents the object detection results on RGB remote sensing datasets. ViTP consistently achieves state-of-the-art performance across DIOR, DIOR-R, and DOTA-v2.0. 
For DOTA-v2.0, which is particularly challenging due to a large quantity of small objects and dense scenes, ViTP achieves a new state-of-the-art of 60.23 mAP. This significantly surpasses previous top performers like BillionFM (58.69), highlighting ViTP's superior ability to handle complex spatial relationships and arbitrary object orientations. To be noticed, SkySense~\cite{skysense} pretraining requires significantly more computational resources, over $17\times$ the GPU hours of ViTP, ViTP still outperforms SkySense. These results underscore the effectiveness of ViTP's instruction-following pretraining in learning robust and task-relevant visual representations for optical remote sensing imagery.

Table~\ref{tab:det_sar_results} presents the object detection results on Synthetic Aperture Radar (SAR) datasets. SAR imagery poses unique challenges due to speckle noise, different scattering mechanisms, and a lack of visual texture compared to optical images. 
On SARDet-100K, ViTP achieves 59.7 mAP, significantly outperforming the previous state-of-the-art SARATR-X (57.3) and other detectors. For RSAR, ViTP sets a new state-of-the-art with 72.31 mAP. 
These remarkable improvements on SAR datasets demonstrate ViTP's strong generalization capabilities and its effectiveness in handling challenging modalities. 

\begin{table}[!t]
\caption{Semantic segmentation (mIoU \%) on optical datasets.}
\label{tab:seg_results}
\renewcommand\arraystretch{1.1}  
\setlength{\tabcolsep}{4pt}
  \centering \footnotesize
\begin{tabular}{c|cc||cc}
model & iSAID  & LoveDA & model & UAVid\\  \Xhline{1pt}

SeCo~\cite{seco} & 57.20  & 43.63 & CANet~\cite{CANet} & 63.50 \\
DenseCLIP~\cite{denseclip} & 59.23  & 49.58 & MP-Former~\cite{mpformer} & 63.67 \\
SatMAE~\cite{satmae} & 62.97 & -&  ABCNet~\cite{ABCNet} & 63.80\\
CACo~\cite{caco} & 64.32 & 48.89 & DecoupleNet~\cite{DecoupleNet} & 65.80 \\ 
RVSA~\cite{rvsa} & 64.49 & 52.44&  CoaT~\cite{CoaT} & 65.80 \\
RSSFormer~\cite{RSSFormer} &  65.55 & 52.43 &  UNetFormer~\cite{UNetFormer} & 67.80 \\
ScaleMAE~\cite{scalemae} & 65.77 & - & MaskFormer~\cite{maskformer} & 68.54 \\
GASSL~\cite{gassl} & 65.95  & 48.76 & LSKNet~\cite{lsknetijcv} & 70.00 \\ 
CMID~\cite{cmid} & 66.21 & - & Segmenter~\cite{Segmenter} & 70.20 \\
TOV~\cite{tov} & 66.24 & 49.70 &  RSSFormer~\cite{RSSFormer} &  70.69 \\ 
RingMo~\cite{RinMo} & 67.20 & - &  DeepLabv3+~\cite{deeplabv3+} & 71.33 \\
SatLas~\cite{satlas} & 68.71 & -& SegFormer~\cite{SegFormer} & 71.44  \\
Sel-MAE~\cite{selectivemae} & - & 53.92 & DenseCLIP~\cite{denseclip} & 71.54 \\
LSKNet~\cite{lsknet} & - & \underline{54.00} & PSPNet~\cite{PSPNet} & 71.71 \\
SkySense~\cite{skysense} & \underline{70.91} & - & OCRNet~\cite{ocrnet} & \underline{71.84} \\ \hline
\textbf{ViTP} & \textbf{71.14}  & \textbf{54.28}& \textbf{ViTP} & \textbf{73.39}
\end{tabular}
\end{table}

\begin{table}[!t]
\caption{Detailed object detection and instance segmentation performance on the SSDD Dataset (SAR modality).}
\label{tab:ssdd_results}
\renewcommand\arraystretch{1.1}  
\setlength{\tabcolsep}{1pt}
  \centering \footnotesize
\begin{tabular}{c|ccc|ccc} 
Model & $AP_{box}$ & $AP_{box}^{75}$ & $AP_{box}^{75}$  & $AP_{mask}$ & $AP_{mask}^{50}$  & $AP_{mask}^{75}$ \\ [2pt]  \Xhline{1pt} 
BoxInst~\cite{boxinst}          & 44.76&   83.75  & 44.11  & 34.10&  71.16  &  27.27  \\
Mask2Former~\cite{mask2former}  & 53.40&   78.45  & 67.02  & 56.52&  85.10  &  69.48  \\
InstaBoost~\cite{instaboost}    & 54.77&   87.85  & 58.54  & 58.95&  89.05  &  71.57  \\
CondInst~\cite{condinst}        & 57.89&   92.53  & 67.40  & 50.31&  90.46  &  54.80  \\
SAM-Seg~\cite{sam}              & 62.41&   94.32  & 75.38  & 59.46&  92.79  &  72.17  \\
CATNet~\cite{catnet}            & 64.66&   \underline{96.46}  & 79.81  & 64.11&  \underline{96.35}  &  77.87  \\ 
HQ-ISNet~\cite{hqisnet}         & 65.58&   95.48  & 80.76  & \underline{64.75}&  95.26  &  \underline{81.70}  \\
RSP-Query~\cite{rsprompter}    & 66.50&   95.80  & 81.81  & 64.57&  95.97  &  81.67  \\ 
SCNet~\cite{scnet}              & \underline{67.25}&   95.75  & \underline{83.38}  & 62.66&  94.75  &  76.53  \\ \hline
\textbf{ViTP}                   & \textbf{70.80}& \textbf{97.80}  & \textbf{86.60}& \textbf{65.90} & \textbf{96.80} & \textbf{81.80}
\end{tabular}
\end{table}

\subsubsection{\textbf{Semantic Segmentation}}

Table~\ref{tab:seg_results} and Table~\ref{tab:ssdd_results} demonstrate ViTP's strong performance in remote sensing semantic segmentation. It sets new state-of-the-art performance on iSAID, LoveDA, UAVid, and SSDD datasets. Especially for UAVid and SSDD datasets, ViTP surpasses previous state-of-the-art by a large margin.
The consistent improvements across these semantic segmentation benchmarks confirm that ViTP's instruction-following pretraining on complex fine-grained regional understanding questions enables the ViT backbone to capture rich semantic information, which is crucial for accurate pixel-level classification.

\begin{table}[!t]
\caption{Change detection performance (F1-Score \%) on SVCD, WHU, LEVER-CD and S2Looking datasets.}
\label{tab:cd_results}
\renewcommand\arraystretch{1.1}  
\setlength{\tabcolsep}{4pt}
  \centering \footnotesize
\begin{tabular}{ccccc}
\multicolumn{1}{c}{\textbf{Model}} & \multicolumn{1}{c}{\textbf{SVCD}}& \multicolumn{1}{c}{\textbf{WHU}}& \multicolumn{1}{c}{\textbf{LEVIR}}  & \multicolumn{1}{c}{\textbf{S2Looking}} \\ \Xhline{1pt}

Scale-MAE~\cite{scalemae}   & - & - & 86.60 & 50.20 \\
SeCo~\cite{seco}   & - & - & 88.40 & 66.00 \\
CACo~\cite{caco}   & - & - & 89.20 & 65.90 \\
GASSL~\cite{gassl}   & - & - & 89.60 & 66.30 \\
SatMAE~\cite{satmae}   & - & - & 90.00 & 65.00 \\
SatMAE++~\cite{satmae++}   & - & - & 90.70 & 56.40 \\
CGNet~\cite{CGNet} & - & - & 92.01  & 64.33  \\
Changer~\cite{changer} & - & - &  92.06 &  67.08 \\
DiFormer~\cite{lin2024diformer} & - & - &  92.15 & 66.31  \\
Changen2~\cite{changen2}   & - & - & 92.20 & \underline{69.10} \\
SkySense~\cite{skysense} & - & - &  \underline{92.58} &  - \\
CLNet~\cite{clnet}   & 92.10  & - & 90.00 &  - \\
SRCDNet~\cite{srcdnet}   & 92.94  & 87.40  & - & -  \\
ESCNet~\cite{escnet}   & 93.54  & - & - & -  \\
DSAMNet~\cite{DSAMNet}   & 93.69  &  - & - &  - \\
GCD-DDPM~\cite{gcdddpm}   & 94.93  &  92.54 & 90.96 &  -  \\
CDContrast~\cite{cdcontrast}   & 95.11  & -  & - &  - \\
DDPM-CD~\cite{ddpmcd}  & 95.62  &  92.65  & 90.91 &  -  \\
DMNet~\cite{DMNet}  & 95.93  & - & - & - \\
SNUnet~\cite{snunet}  & 96.20   &  83.49 & 88.59 & 63.19 \\
BIT~\cite{bit}  & -   &  83.98 & 89.31 &  63.76   \\
BiFA~\cite{snunet}  & -  & 94.37  &  90.69 & - \\
SGSLN~\cite{sgsln}   & 96.24  & 94.67 & 91.93&  -  \\
RSP~\cite{rsp} & 96.81  & - & 90.93 &  - \\
SAAN~\cite{saan}  & 97.03  & - & 91.41 & - \\
SiamixFormer~\cite{siamixformer}  & 97.13  & - & 91.58 & - \\
TransUNetCD~\cite{transunetcd}  &  97.17 & 93.59 & 91.11 &  -  \\
RDPNet~\cite{rdpnet}  & 97.20  & - &91.20 &  -\\
SDACD~\cite{sdacd}   & 97.34  &  & &    \\
WNet~\cite{wnet}   & 97.56  & 91.25 & 90.67 &   - \\
ChangeMamba~\cite{changemamba} & -  & 92.55 & 90.16 & -  \\
RS-Mamba~\cite{rsmamba}  &  - &  92.79 &  89.77 &  -  \\
ChangeFormer~\cite{changeformer} & - & 93.04 &  91.11 & 63.39  \\
CDMamba~\cite{cdmamba} &  -  & 93.76 &  90.75 & 67.08 \\
LSKNet~\cite{lsknetijcv} & - & 92.06 & 92.27 & 67.52 \\
RVSA~\cite{rvsa}  & 97.78  & 94.07 & 92.52 &  - \\
ChangeCLIP~\cite{changeclip}   & 97.89  & \underline{94.82} & 92.01 & - \\
P2V-CD~\cite{p2vcd}   & \underline{98.42}  & 92.38 & 91.94 & - \\ \hline
\textbf{ViTP}  & \textbf{98.63}  & \textbf{94.98 } & \textbf{92.67} &   \textbf{69.89}   \\
\end{tabular}
\end{table}

\subsubsection{\textbf{Change Detection}}

The comprehensive results, presented in Table~\ref{tab:cd_results}, show that ViTP achieves state-of-the-art performance across all of SVCD, WHU, LEVIR-CD and S2Looking datasets. For the SVCD dataset, which contains more diverse change types, ViTP shows a clear performance advantage. On the LEVIR-CD dataset, which focuses on building changes, ViTP achieves a new state-of-the-art F1 of 92.67. This surpasses previous methods, demonstrating its proficiency in identifying fine-grained changes in complex suburban environments. Similarly, on the WHU-CD dataset, ViTP sets a new benchmark with an F1 of 94.98, effectively handling variations in building scale and appearance. These significant improvements underscore the rich semantic understanding power of ViTP for temporal analysis.

\subsection{Medical Imaging}

Table~\ref{tab:med_seg_results} demonstrates the strong performance of ViTP in medical semantic segmentation. ViTP achieves new state-of-the-art results on the AMOS, BraTS, and ConvidQUEx datasets. Specifically, similar to ViTP, SAM-based approaches (MedSAM~\cite{medsam}, SAMMed2D~\cite{sammed2d} and IMIS-SAM~\cite{imissam}) are also finetuned on the downstream datasets. However, ViTP achieves substantial performance gains, which can be attributed to its more effective pretraining strategy. When compared to the classic medical segmentation model (nnU-Net~\cite{nnunet}), ViTP not only exhibits superior generalization but also surpasses its performance, despite nnU-Net being a highly specialized model. Finally, a comparison with MedMAE~\cite{medmae}, a recent 2D medical pretraining method, highlights the effectiveness of our proposed pretraining paradigm.

\begin{table}[!t]
\caption{Semantic segmentation performance (mDice \%) on Medical Imagery Datasets.}
\label{tab:med_seg_results}
\renewcommand\arraystretch{1.1}  
\setlength{\tabcolsep}{6pt}    
  \centering \footnotesize       
\begin{tabular}{c|ccc} 
model & AMOS2022  & BraTS2021 & CovidQUEx \\ 
\Xhline{1pt}  
\textit{5-points Prompt} &&&\\
MedSAM~\cite{medsam} & 81.36  & 69.44 & 78.18 \\
SAMMed2D~\cite{sammed2d} & 87.81 & 70.70 & 83.39 \\ 
IMIS-SAM~\cite{imissam} & 87.42 & 64.77 & 86.28 \\ 
\hline
\textit{Box Prompt} &&&\\
MedSAM~\cite{medsam} & 86.74  & 71.23 & 78.87 \\
SAMMed2D~\cite{sammed2d} & 88.67 & \underline{71.26} & 77.81 \\ 
IMIS-SAM~\cite{imissam} & 88.71 & 70.59 & 82.91 \\ 
\hline
nnUNet~\cite{nnunet} & 87.28  & 71.03 & \underline{90.41} \\
MedMAE~\cite{medmae} & \underline{90.11} & 69.59 & 90.18 \\ 
\hline
\textbf{ViTP} & \textbf{90.60}  & \textbf{72.11} & \textbf{91.55} \\  
\end{tabular}
\end{table}

\section{Analysis and Ablation Study}
To comprehensively understand the impact of various components and hyperparameters on ViTP's performance, we conduct a series of rigorous ablation studies and analytical experiments. This section details the ablation studies, including: (1) validating the principles of our data recipe, (2) assessing the impact of pre-training steps, (3) evaluating the effect of the VRL image token dropping ratio, and (4) analyzing the influence of the ViTP language model's size.

For these studies, we utilize the ViTP model pre-trained on remote sensing data and evaluate its performance on the RSAR dataset, a challenging benchmark for oriented object detection in SAR imagery. We selected this specific downstream task for several reasons. Firstly, the SAR modality presents unique challenges not found in optical imagery, making it an excellent testbed for a model's ability to generalize to specialized visual domains. Secondly, oriented object detection demands precise localization, which stringently tests the model's capacity to capture fine-grained spatial information and recognize weak patterns amidst dense objects. Therefore, performance on the RSAR dataset serves as a fair and insightful proxy for the model's overall representational power. For experimental efficiency, unless otherwise specified, models are finetuned on the RSAR validation set and evaluated on its test set.

Furthermore, we investigate the unique advantages of the ViTP paradigm over other pre-training methods during the finetuning stage, specifically focusing on its data efficiency and robustness to various data corruptions.

\begin{table}[!t]
\caption{Ablation of the data recipe. Performance on RSAR confirms the necessity of our comprehensive data curation strategy.}
\label{tab:data_recipe}
\renewcommand\arraystretch{1.1}  
\setlength{\tabcolsep}{6pt}
  \centering \footnotesize
\begin{tabular}{c|c}
Pretrain Paradigm & RSAR mAP  \\  \Xhline{1pt}
w/o Diversity &  52.6 \\
w/o SAR &  52.5 \\
w/o Grounding & 53.0 \\
w/o General data & 52.3 \\ \hline
full data & \textbf{54.6}  \\ 
\end{tabular}
\vspace{-8pt}
\end{table}

\subsection{Data Recipe}
Our ViTP pretraining dataset was carefully curated according to the principles stated in Section~\ref{sec:dataset_recipe}. Table~\ref{tab:data_recipe} ablates the contribution of different data types, where ``w/o Diversity'' refers to the exclusion of diverse remote sensing visual instruction datasets (Million-AID, GAIA, Levir-CC, VHM, VRSBench, and RSVG) during pre-training. Removing any single component: diverse datasets, SAR-specific datasets, grounding datasets, or general-domain datasets, leads to a significant drop in downstream performance on the RSAR dataset. Excluding SAR data prevents the model from learning modality-specific features, while removing grounding data impairs its localization ability. Omitting general data leads to overfitting on specialized patterns, which leads to the most severe performance drop. This confirms that a diverse mixture of diverse, general, domain-specific, and task-specific instruction data is crucial for optimal performance.

\subsection{Impact of Pretraining Steps}

\begin{figure}[!t]
\centering
\includegraphics[width=0.85\linewidth]{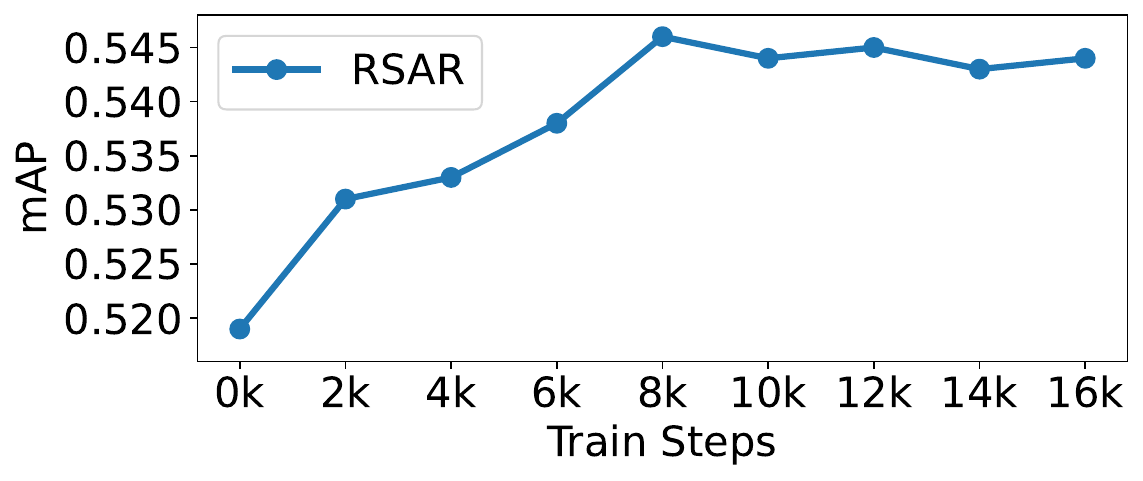}
\vspace{-4pt}
\caption{Effect of Pretraining Duration. RSAR mAP improves with more pretraining steps before saturating at $\sim$8k steps.}
\label{fig:abl_steps}
\end{figure}

The duration of pre-training is a significant factor influencing downstream task performance. As depicted in Figure~\ref{fig:abl_steps}, the model's mAP on the RSAR dataset generally improves with an increasing number of pre-training steps. This trend indicates that more extensive instruction tuning enables the model to learn progressively better feature representations. However, the performance curve begins to plateau around 8k steps. Consequently, we set the number of ViTP pre-training steps to 8k by default for our main model, balancing performance with computational cost. This pre-training phase takes approximately 23 hours to complete 8k steps training on $8\times$ A40 GPUs, demonstrating the efficiency of our proposed method.

\subsection{Impact of VRL Token drop Ratio}
The Visual Robustness Learning (VRL), introduced in Section~\ref{sec:vrl}, involves randomly dropping a portion of image tokens during pre-training. This study investigates the optimal dropping ratio for this process. An appropriate ratio compels the ViT to learn more robust and semantically rich representations by inferring missing visual context. Conversely, an excessively high ratio risks obscuring essential image information, thereby posing an ill-posed training target and hindering effective learning.

\begin{figure}[!t]
    \centering
    \begin{minipage}[t]{0.52\linewidth}
        \centering
        \includegraphics[width=\linewidth]{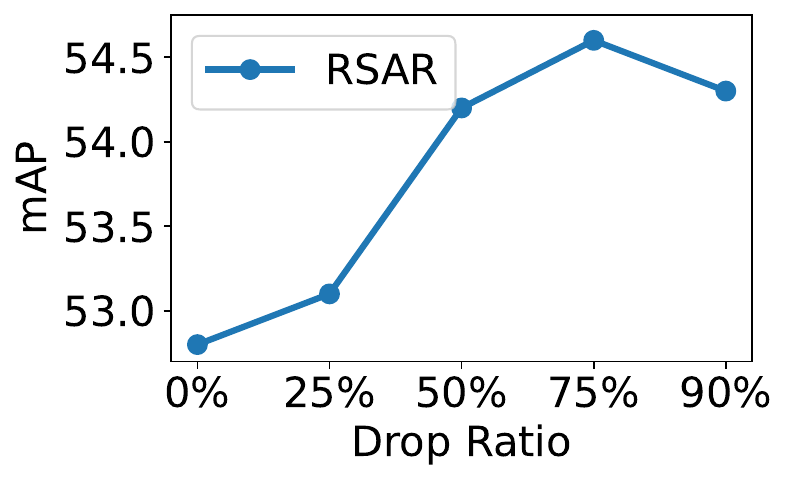}
        \caption{Impact of the VRL Drop Ratio. ViTP performance on RSAR reaches a peak with a 75\% token drop ratio.} 
        \label{fig:abl_mask}
    \end{minipage}
    \hfill
    \begin{minipage}[t]{0.42\linewidth}
        \centering
        \includegraphics[width=\linewidth]{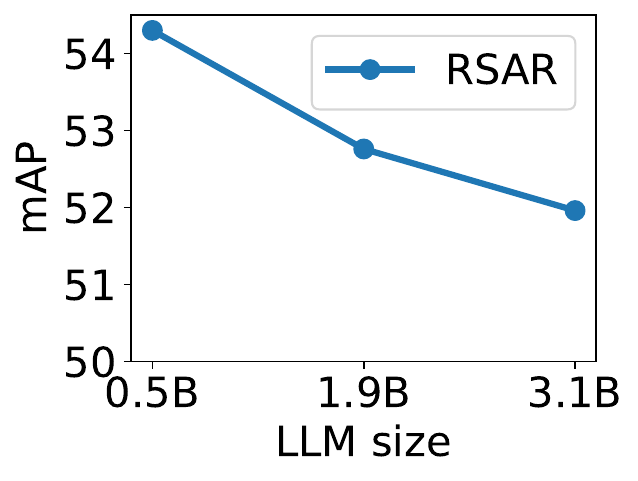}
        \caption{Impact of LLM size. A larger LLM hinders the performance of ViTP on the RSAR Benchmark.}
        \label{fig:abl_llm}
    \end{minipage}
\end{figure}

As shown in Figure~\ref{fig:abl_mask}, the model's performance on the RSAR dataset varies significantly with different dropping ratios. The performance peaks at a ratio of 75\%, which improves the mAP from 52.8 (without VRL) to 54.6. This suggests that a 75\% masking ratio effectively regularizes the model without excessively impeding information flow. Ratios from 0\% to 50\% appear to provide insufficient regularization. Surprisingly, even with a very high masking rate of 90\%, the model still benefits from this strong regularization, achieving a notable 54.3 mAP.

\subsection{Impact of Language Model Size}
This study investigates the influence of the LLM's capacity on ViTP's overall performance. As shown in Figure~\ref{fig:abl_llm}, as the LLM size increases, the downstream performance on the RSAR dataset steadily degrades.
This phenomenon is analogous to the findings in Masked Autoencoders~\cite{mae} pretrain paradigm, which highlight the importance of a lightweight decoder. It indicates that for the challenging SAR modality, a more powerful LLM might compensate for suboptimal visual features extracted by the ViT. This could reduce the optimization pressure on the ViT, preventing it from learning the most effective modality-specific representations.

\begin{table}[!t]
\caption{Pretraining efficiency and performance. Time is estimated for pretraining on $8\times$ A40 GPUs. ViTP achieves SOTA or competitive performance with a fraction of the computational cost.}
\label{tab:train_time}
\renewcommand\arraystretch{1.1}  
\setlength{\tabcolsep}{6pt}
  \centering \footnotesize
\begin{tabular}{cc|ccc}
 & Methods & Hours & DIOR-R& iSAID\\  \Xhline{1pt}
\multirow{2}{*}{\begin{tabular}[c]{@{}c@{}}MIM \end{tabular}} & RVSA~\cite{rvsa} & 250 & 70.96& 64.49\\
 & Scale-MAE~\cite{scalemae} & \underline{60} & 70.20 & 65.77\\ \hline
\multirow{2}{*}{\begin{tabular}[c]{@{}c@{}}CL \end{tabular}} & RemoteCLIP~\cite{remoteclip} & 100 & 70.20& 62.53\\
 & SkySense~\cite{skysense} & 400 & \underline{74.27}& \underline{70.91} \\ \hline
 & \textbf{ViTP} &  \textbf{23} & \textbf{75.08}& \textbf{71.14} \\ 
\end{tabular}
\end{table}

\subsection{Training Efficiency}

A key advantage of the ViTP paradigm is its pretraining efficiency, which lowers the barrier for developing specialized foundation models. As detailed in Table~\ref{tab:train_time}, ViTP pretraining completes in one day on 8 A40 GPUs. This is dramatically faster than competing paradigms and achieves better downstream performance. For instance, ViTP is \textbf{2.6x} faster than the efficient MIM-based method (Scale-MAE) and over \textbf{17x} faster than the high-performing but computationally intensive contrastive learning method, SkySense.

\section{Conclusion}
This paper addressed the underexplored, top-down pathway from high-level understanding to low-level perception in vision foundation models. We introduced Visual insTruction Pretraining (ViTP), a novel paradigm that leverages the high-level semantic understanding of a Vision-Language Model to directly guide the feature learning of a ViT backbone. Complemented by our Visual Robustness Learning (VRL), ViTP compels the vision model to learn rich, robust, and task-relevant representations from downstream-specific instructional data. Our extensive experiments on 16 challenging remote sensing and medical imaging benchmarks validated the effectiveness of the proposed ViTP method. ViTP not only establishes new SOTA performance across a diverse range of downstream tasks but also proves remarkably computationally efficient. We believe this work opens a promising new avenue for a deeper integration of high-level understanding into the core of visual feature learning.


\appendix

\section{More Related Work}
\subsection{Remote sensing Perception Foundation Model}
In the remote sensing domain, pretraining strategies are tailored to address unique challenges such as large scale variations, multi-spectral imagery, and domain-specific object categories~\cite{skysense,scalemae,li2024predicting,su2023lightweight}. Supervised methods like SAMRS~\cite{samrs} and MSFA~\cite{sardet} achieve strong performance by leveraging annotated remote sensing datasets. To exploit the abundance of unlabeled data, MIM-based approaches such as RingMo~\cite{RinMo}, SatMAE~\cite{satmae}, and Scale-MAE~\cite{scalemae} are proposed, specifically designed to model the dense small objects, multi-modal or multi-scale characteristics of satellite imagery. Meanwhile, contrastive learning has been explored through both image-text and image-image formulations. Models like GeoRSCLIP~\cite{rs5m} and RemoteCLIP~\cite{remoteclip} leverage large-scale image-text pairs to enable zero-shot classification capabilities. Image-based contrastive methods such as CACo~\cite{caco} and Skysense~\cite{skysense} focus on learning discriminative features from unlabeled remote sensing imagery. Hybrid approaches like CMID~\cite{cmid} and GFM~\cite{gfm} further integrate the strengths of both MIM and contrastive learning to learn more comprehensive and robust representations.

\subsection{Medical Images Perception Foundation Model}
Pretraining vision foundation models is particularly crucial in medical imaging, where labeled data is scarce and expensive to obtain. A prominent line of research focuses on adapting the general-purpose Segment Anything Model (SAM)~\cite{sam} to the medical domain. Adaptations range from finetuning SAM on large-scale medical datasets, as in MedSAM~\cite{medsam}, to pretraining SAM-like architectures from scratch on medical images, as demonstrated by SAM-Med2D~\cite{sammed2d}. Other frameworks, such as IMIS-SAM~\cite{imissam}, aim to enhance the interactive capabilities of SAM for clinical workflows. Concurrently, specialized MIM pretraining methods like MedMAE~\cite{medmae} and S3D~\cite{s3d} are developed to better capture the unique characteristics of medical data.

A common limitation across these domain-specific foundation models is their heavy reliance on tailored architectural or training designs, which restricts their transferability to other domains. In contrast, our proposed ViTP framework offers a more generalizable pretraining strategy that can be seamlessly adapted to various downstream domains by simply curating the corresponding visual instruction datasets.

\subsection{Continual Pretraining} 
Continual pretraining was first popularized in NLP, where Gururangan et al.~\cite{dontstop} demonstrated that continuing pretraining on in-domain data significantly improves model performance. This concept is also successfully adapted to computer vision. For instance, CSPT~\cite{zhang2022consecutive} first pretrains on ImageNet and then continues pretraining on a target remote sensing dataset using an MIM objective. TOV~\cite{tov} adopts a curriculum strategy, freezing early layers pretrained on natural images while finetuning deeper layers on specialized data. To reduce computational overhead, RemoteCLIP~\cite{remoteclip} initializes from pretrained CLIP~\cite{clip} weights and continues pretraining on domain-specific data. Similarly, MSFA~\cite{sardet} adopts a multi-stage strategy: pretraining on ImageNet, followed by training on optical remote sensing detection datasets, and finally finetuning on SAR detection data.

Inspired by these successes, we adopt a continual pretraining approach, initializing our VLM from the well-established InternVL~\cite{internvl25} model. However, our method differs from prior work in two key aspects: (1) to mitigate catastrophic forgetting of general visual knowledge, we incorporate a small proportion of general-domain data into the specialized training corpus; (2) we introduce Visual Robustness Learning as an additional regularizer to enhance the model's robustness during continual pretraining.

\section{Implementation Details}

Training a VLM from scratch conventionally requires a multi-stage pipeline: vision–language contrastive pretraining, projector alignment, and large-scale instruction tuning. To bypass the prohibitive cost of these initial phases, we instead bootstrap from a publicly available, high-capacity VLM. Specifically, we initialise our weights from InternVL-2.5~\cite{internvl25}, which uses a custmized ViT-Large backbone~\cite{internvl} and a Qwen2~\cite{qwen} language model. All pretraining are performed on 8× NVIDIA A40 (48 GB) GPUs with a global batch size of 128, and all downstream task finetunings are conducted on $8\times$ NVIDIA RTX3090 (24 GB) GPUs. For ViTP pretraining, we use the AdamW optimizer with a learning rate of 2e-5 and a cosine decay schedule. For remote sensing and medical image foundation models, we apply a schedual of 8,000 training steps, however for the general domain foundation model, we pretrain for 16,000 steps. To accelerate both pre-training and downstream finetuning, Flash-Attention~\cite{flashattention} is integrated into every Vision Transformer self-attention layer, yielding extra throughput gain without altering convergence behaviour.

\section{Experimental Datasets }

\subsection{Pretraining Datasets}
For the ViTP pretraining phase, we leverage a diverse set of publicly available image-text paired instruction datasets, adhering to the principles outlined in Section ``\textit{Pretraining Dataset Recipe}''. 

\subsubsection{General Domain}
The pretraining datasets are listed in Table~\ref{recipe_general}. In the dataset recipe, ShareGPT4V~\cite{sharegpt4v}, DVQA~\cite{dvqa}, ChartQA~\cite{chartqa}, AI2D~\cite{ai2d}, DocVQA~\cite{docvqa}, and GeoQA+~\cite{GeoQA} are diverse visual question answering (VQA) datasets, providing rich instructional data for complex reasoning, chart and document understanding, and geospatial question answering. The SynthDoG-EN~\cite{SynthDoG} dataset specifically addresses the unique challenges of Optical Character Recognition (OCR) tasks. ShareGPT4V~\cite{sharegpt4v} also provides visual grounding (VG) data, enhancing spatial localization capabilities. GAIA~\cite{zavras2025gaia} and Million-AID~\cite{millionaid} contribute foundational captioning (Caption) and classification (CLS) tasks, ensuring the model develops robust general-purpose vision-language understanding. This comprehensive pretraining dataset recipe enables the model to learn robust, instruction-aware visual representations that are highly relevant to diverse general domain tasks.

\begin{table}[!t]
\caption{Composition of the pretraining dataset for general domain.}
\renewcommand\arraystretch{1.1}  
\setlength{\tabcolsep}{2pt}
\centering \footnotesize
\begin{tabular}{l|ccc}
~~~~~~Dataset & Size & Sample Rate & Tasks \\ \Xhline{1pt} 
ShareGPT4V~\cite{sharegpt4v} & 767k & 1 & VQA,VG \\ 
DVQA~\cite{dvqa} & 200k & 1 & VQA \\
ChartQA~\cite{chartqa} & 18k & 1 & VQA \\
AI2D~\cite{ai2d} & 12k & 1 & VQA \\
DocVQA~\cite{docvqa} & 10k & 1 & VQA \\
GeoQA+~\cite{GeoQA} & 72k & 1 & VQA \\
SynthDoG-EN~\cite{SynthDoG} & 30k & 1 & OCR \\
GAIA~\cite{zavras2025gaia} & 33K & 0.2 & Caption \\
Million-AID~\cite{millionaid} & 920k & 0.01 & Caption,CLS \\ 
\end{tabular}
\label{recipe_general}
\end{table}

\subsubsection{Remote Sensing Domain}
The pretraining datasets are listed in Table~\ref{recipe_rs}. In the dataset recipe, Million-AID~\cite{millionaid}, GAIA~\cite{zavras2025gaia}, LevirCCcaptions~\cite{LevirCC}, VHM~\cite{vhm}, RSVQA~\cite{rsvqa}, and FIT\_RS~\cite{skysensegpt} are large-scale remote sensing visual instruction datasets, providing diverse remote sensing scenes and visual tasks. ISPRS\_SAR~\cite{isprs_sar} and SAR Sentinel-1\&2~\cite{sentinel12_image_pairs} datasets specifically address the unique challenges of Synthetic Aperture Radar (SAR) data. We construct question-answer pairs related to object identification, scene interpretation, and attribute recognition in SAR images using these existing SAR classification datasets. This ensures modality alignment and task relevance for SAR-specific downstream applications. GeoChat~\cite{geochat}, DIOR-RSVG~\cite{diorrsvg}, RSVG~\cite{rsvg}, and VRSBench~\cite{vrsbench} datasets primarily focus on visual grounding tasks within the remote sensing domain, providing rich instructional guidance for spatial understanding, object attribute recognition, and target localization. The mini-InternVL~\cite{miniinternvl} dataset contributes to general vision-language understanding capabilities, ensuring that ViTP retains broad applicability and does not overfit to highly specialized remote sensing patterns. The comprehensive pretraining dataset recipe enables ViTP to learn robust, instruction-aware visual representations that are highly relevant to diverse remote sensing tasks. 
\begin{table}[!t]
\caption{Composition of the pretraining dataset for remote sensing domain.}
\renewcommand\arraystretch{1.1}  
\setlength{\tabcolsep}{2pt}
\centering \footnotesize
\begin{tabular}{l|ccc}
~~~~~~Dataset & Size & Sample Rate & Tasks \\ \Xhline{1pt}
Mini-InternVL~\cite{miniinternvl} & 1394k & 0.03 & Caption,VQA,OCR \\
RSVQA~\cite{rsvqa} & 100k & 0.1 & VQA \\
FIT\_RS~\cite{skysensegpt} & 100k & 0.1 & VQA \\
GeoChat~\cite{geochat} & 64k & 2 & VG \\
VRSBench~\cite{vrsbench} & 38k & 5 & VG \\
RSVG~\cite{rsvg} & 5.5k & 10 & VG \\
DIOR-RSVG~\cite{diorrsvg} & 27k & 8 & VG \\
ISPRS\_SAR~\cite{isprs_sar} & 1.5k & 1 & CLS \\
SAR\_Sentinel-1\&2~\cite{sentinel12_image_pairs} & 16k & 1 & CLS \\
VHM~\cite{vhm} & 223k & 1 & Caption,VQA,CLS \\
LevirCCcaptions~\cite{LevirCC} & 50k & 0.5 & Caption \\
GAIA~\cite{zavras2025gaia} & 33k & 1 & Caption \\
Million-AID~\cite{millionaid} & 920k & 0.05 & Caption,CLS
\end{tabular}
\label{recipe_rs}
\end{table}

\begin{table}[!t]
\caption{Composition of the pretraining dataset for medical imagery domain. }
\renewcommand\arraystretch{1.1}  
\setlength{\tabcolsep}{2.5pt}
\centering \footnotesize
\begin{tabular}{l|ccc}
~~~~~~Dataset & Size & Sample Rate & Tasks \\ \Xhline{1pt}
Mini-InternVL~\cite{miniinternvl} & 1394k & 0.02 & Caption,VQA,OCR \\
GMAI-MMBench~\cite{ye2024gmai} & 5k & 20 & VQA \\
Open-i~\cite{openi} & 7k & 5 & Caption \\
Huatuo-OA~\cite{huatuo} & 647k & 0.1 & Caption \\
Huatuo-VQA~\cite{huatuo} & 647k & 0.5 & VQA \\
PMC-OA~\cite{lin2023pmc_oa} & 1647k & 0.05 & Caption \\
PMC-VQA~\cite{zhang2023pmc_vqa} & 227k & 0.5 & VQA \\
OmniMedical~\cite{hu2024omnimedvqa} & 89k & 1 & VQA \\
Quilt-1M~\cite{ikezogwo2023quilt_1m} & 723k & 0.1 & Caption \\
Quilt-Instruct~\cite{seyfioglu2024quilt_i} & 147k & 1 & VQA \\
\end{tabular}
\label{tab:medical_recipe}
\end{table}

\subsubsection{Medical Imagery Domain}
Similar to the approach for remote sensing, we collected several open-source medical image-text paired instruction datasets.
The pretraining datasets are listed in Table~\ref{tab:medical_recipe}. Among them, Huatuo~\cite{huatuo}, PMC~\cite{lin2023pmc_oa}, and OmniMedical~\cite{hu2024omnimedvqa} are large-scale, general medical visual instruction datasets covering various medical imaging modalities, including CT, MRI, ultrasound, and others. These datasets reframe various downstream tasks into a question-answering format, augmented with millions of descriptive captions, enabling the model to build comprehensive knowledge across diverse medical imaging modalities. Moreover, since medical image segmentation is a key downstream task, we also include the GAMI-MMBench~\cite{ye2024gmai} dataset. It provides VQA based on ROIs and segmentation masks, which aids in aligning the model with dense prediction tasks. The Open-i~\cite{openi} and Quilt~\cite{ikezogwo2023quilt_1m} datasets are included to compensate for the relative scarcity of X-ray and pathology images in other general medical collections. Finally, as with the remote sensing domain, mini-InternVL~\cite{miniinternvl} is incorporated to enhance general vision-language understanding.
 
\subsection{Domain-Specific Finetuning Datasets}

\begin{table}[!t]
\caption{Overview of downstream task datasets.}
\label{tab:det_datasets}
\renewcommand\arraystretch{1.1}  
\setlength{\tabcolsep}{1pt}
\centering \footnotesize
\begin{tabular}{cc|cccc}
Task & Dataset & Modal & Box Format & Images & Classes \\  \Xhline{1pt}
\textit{\textbf{General Domain} }\\  \Xhline{1pt}
\multirow{3}{*}{\begin{tabular}[c]{@{}c@{}}Detection and \\ Segmentation \end{tabular}} & COCO~\cite{coco} & RGB &  Polygons &  118,287 & 80 \\
 & COCO~\cite{coco} & RGB & Hori. Box &  118,287  & 80 \\
 & ADE20K~\cite{ade20k} & RGB & Mask & 20,210 & 150 \\ \Xhline{1pt}
\textit{\textbf{Remote Sensing} }\\  \Xhline{1pt}
\multirow{6}{*}{\begin{tabular}[c]{@{}c@{}}Object\\ Detection\end{tabular}} & DIOR~\cite{dior} & RGB & Hori. Box & 23,463 & 20 \\
 & DIOR-R~\cite{diorr} & RGB & Ori. Box & 23,463 & 20 \\
 & DOTA-v2.0~\cite{dota} & RGB & Ori. Box & 11,268 & 18 \\
 & SARDet-100K~\cite{sardet} & SAR & Hori. Box & 116,598 & 6 \\
 & SSDD~\cite{ssdd} & SAR & Hori. Box & 1,160 & 1 \\ 
 & RSAR~\cite{rsar} & SAR & Ori. Box & 95,842 & 6 \\ \hline
\multirow{4}{*}{\begin{tabular}[c]{@{}c@{}}Semantic and \\ Instance \\  Segmentation\end{tabular}} & iSAID~\cite{isaid} & RGB &  Mask & 2,806 & 15 \\
 & LoveDA~\cite{loveda} & RGB & Mask & 5,987 & 7 \\
 & UAVid~\cite{UAVid} & RGB & Mask & 5,510 & 8 \\
 & SSDD~\cite{ssdd} & SAR & Polygons & 1,160 & 1 \\ \hline
\multirow{4}{*}{\begin{tabular}[c]{@{}c@{}}Change \\ Detection \end{tabular}} & SVCD~\cite{SVCD} & RGB &  Mask &  16,000 & 2 \\
 & WHU~\cite{whu} & RGB & Mask & 11,456  & 2 \\
 & LEVIR-CD~\cite{LEVIR} & RGB & Mask & 10,192 & 2 \\
 & S2Looking~\cite{S2Looking} & RGB & Mask &  5,000 & 2 \\  \Xhline{1pt}
\textit{\textbf{Medical Image} }\\  \Xhline{1pt}
 \multirow{3}{*}{\begin{tabular}[c]{@{}c@{}}Medical \\ Segmentation \end{tabular}} & AMOS2022~\cite{ji2022amos} & CT &  Mask &  19,310 & 14 \\
 & BraTS2021~\cite{brats} & MRI & Mask & 76,467  & 3 \\
 & CovidQUEx~\cite{covid_quex} & X-ray & Mask & 5,826 & 2 \\
 
\end{tabular}
\end{table}

\subsubsection{Natural Image Detection and Segmentation}

Object detection and segmentation in natural scenes are fundamental tasks in computer vision, serving as benchmarks for general-purpose visual understanding. Object detection involves localizing and classifying multiple objects within everyday photographs, while segmentation provides pixel-level masks for precise delineation. These tasks are crucial for applications ranging from autonomous driving to augmented reality. We evaluate ViTP's versatility on standard natural image benchmarks using two complementary datasets: COCO~\cite{coco} for object detection and instance segmentation, and ADE20K~\cite{ade20k} for semantic segmentation. Following established protocols~\cite{mae,dinov2}, we adopt Mask R-CNN~\cite{maskrcnn} as the default framework for COCO tasks with a 12-epoch training schedule and standard data augmentations. For ADE20K semantic segmentation, we employ UperNet~\cite{UNet} with a 160k iteration training strategy. We report standard COCO metrics (mAP@0.5:0.95) for detection and segmentation tasks on COCO, and mean Intersection over Union (mIoU) for ADE20K semantic segmentation.

\subsubsection{Remote Sensing Object Detection}

Object detection in remote sensing~\cite{dota,orcnn,sm3det,su2025rapid,11016924} involves identifying and precisely localizing objects of interest (e.g., vehicles, ships, and bridges) within aerial or satellite imagery. This task is crucial for applications such as urban planning and disaster monitoring. Challenges include vast scale variations of objects, arbitrary orientations, dense object distributions, and complex backgrounds. To evaluate the applicability of our proposed model for remote sensing object detection tasks under various scenarios, we conducted experiments on the following datasets: DIOR~\cite{dior}, DIOR-R~\cite{diorr}, DOTA-v2.0~\cite{dota}, SARDet-100K~\cite{sardet}, RSAR~\cite{rsar}, and SSDD~\cite{ssdd}. We use the standard COCO evaluation metrics (mean Average Precision, mAP) to evaluate the performance of the models. Following previous practice~\cite{skysense,rsar,sardet}, we use the oriented R-CNN~\cite{orcnn} as the default detector for oriented object detection tasks and Cascade R-CNN~\cite{cascade_rcnn} for horizontal object detection tasks. 
For training and testing, we resized all optical datasets (DIOR, DIOR-R, and DOTA-v2) to a standard size of 1024$\times$1024 pixels. For the DOTA-v2 dataset specifically, we adopted a single-scale approach, cropping images into 1024$\times$1024 patches with a 200-pixel overlap to handle its original large size. The SAR datasets, SARDet-100K and RSAR, were resized to 800$\times$800 pixels.

\subsubsection{Remote Sensing Semantic Segmentation} 

Semantic segmentation in remote sensing~\cite{isaid,UNetFormer,ssdd} involves classifying each pixel in an image into a predefined category, such as land cover types (e.g., forest, water, urban area) or specific objects. This pixel-level understanding is vital for environmental monitoring, urban planning, and resource management. We evaluated ViTP's performance on segmentation tasks using the following widely-used remote sensing datasets: iSAID~\cite{isaid}, LoveDA~\cite{loveda}, UAVid~\cite{UAVid}, and SSDD~\cite{ssdd}. Following previous practice~\cite{skysense,RinMo}, we use the UperNet~\cite{UNet} as the default segmentor for semantic segmentation tasks (iSAID, LoveDA, and UAVid) and Mask-RCNN~\cite{maskrcnn} for instance segmentation tasks (SSDD). We report the mean Intersection over Union (mIoU) as the evaluation metric for semantic segmentation and Average Precision (AP) for instance segmentation tasks.

\subsubsection{Remote Sensing Change Detection}
Change detection in remote sensing focuses on identifying and characterizing differences in the state of an object or phenomenon by observing it at different times. This bitemporal analysis is fundamental for applications such as urban sprawl monitoring, deforestation tracking, and urbanization analysis. The primary challenges in this task include handling variations in illumination and atmospheric conditions between image acquisitions, precise image co-registration, and distinguishing meaningful semantic changes from irrelevant ones. To assess ViTP's capabilities in this domain, we conduct experiments on three widely-recognized public datasets: SVCD~\cite{SVCD}, LEVIR-CD~\cite{LEVIR} and WHU-CD~\cite{whu}. Following the common practice, we employ a simple Siamese UperNet-based framework~\cite{UNet} as the default detector and report the F1-Score as our primary evaluation metric.

\subsubsection{Medical Imaging Semantic Segmentation}
Semantic segmentation in medical image analysis involves classifying each pixel in an image into a predefined category, such as an organ (e.g., lung, kidney, heart) or a specific lesion. This pixel-level understanding is vital for clinical diagnosis, disease analysis, and patient prognosis. We evaluated ViTP's performance on segmentation tasks using three challenging medical datasets: AMOS2022~\cite{ji2022amos}, BraTS2021~\cite{brats}, and ConvidQUEx~\cite{covid_quex}. These datasets encompass three primary clinical modalities: CT (AMOS2022), MRI (BraTS2021), and chest X-ray (ConvidQUEx), and covering both organ and lesion segmentation scenarios. We use UperNet~\cite{UNet} as the default segmentation head. We report the mean Dice Score (mDice) as the evaluation metric for all segmentation tasks.

\begin{table}[!t]
\caption{Comparison of pretraining paradigms for end-to-end finetuning. Results are shown for object detection ($AP_{box}$) and instance segmentation ($AP_{mask}$) on COCO val using Mask R-CNN (1x schedule), and semantic segmentation (mIoU) on ADE20K using UperNet (160k iterations). All models use a ViT-Large backbone.}
\label{tab:coco_comp}
\renewcommand\arraystretch{1.1}  
\setlength{\tabcolsep}{4pt}
  \centering \footnotesize
\begin{tabular}{cc|ccc}
\multirow{2}{*}{\begin{tabular}[c]{@{}c@{}}Pretrain\\ Paradigm\end{tabular}} & \multirow{2}{*}{Model} & \multicolumn{2}{c}{COCO val} & \multirow{2}{*}{\begin{tabular}[c]{@{}c@{}}ADE20K\\ (mIoU)\end{tabular}} \\
 &  & \multicolumn{1}{c}{(AP$_{box}$)} & (AP$_{mask}$) &  \\ \Xhline{1pt}
\multirow{2}{*}{\begin{tabular}[c]{@{}c@{}} Supervised \end{tabular}}   & IN. Cls.~\cite{vit} & 49.3 & 43.9 & 49.9 \\ 
 & DeiT III ~\cite{deit3} & 48.7 & 41.1 & 52.0  \\  \hline
\multirow{3}{*}{\begin{tabular}[c]{@{}c@{}}Contrastive \\ Learning\end{tabular}} 
 & CLIP~\cite{clip} & 51.3 & - & 52.2 \\
 & MoCov3~\cite{mocov3} & 49.3 & 44.0 & 49.1 \\
 & DINOv2~\cite{dinov2} & 53.4 & 46.8 & 55.0 \\ \hline 
\multirow{3}{*}{\begin{tabular}[c]{@{}c@{}}Masked \\ Image \\ Modelling\end{tabular}}  
& BEiT~\cite{beit} & 53.3 & 47.1 & 53.3 \\
 & MAE~\cite{mae} & 53.3 & \textbf{47.2} & 53.6 \\
 & 4M~\cite{4m} & 53.7 & 46.4 & 53.4 \\
 \hline
Understanding & \textbf{ViTP} & \textbf{53.9} & 46.7 & \textbf{55.8}
\end{tabular}
\end{table}

\section{More Experiments}
\subsection{General Domains}
To validate the efficacy of our proposed ViTP paradigm, we conducted comprehensive experiments against SOTA pretraining methods in two rigorous settings: end-to-end finetuning and frozen-backbone feature evaluation. All comparisons utilize a ViT-Large backbone for fair assessment.

As shown in Table~\ref{tab:coco_comp}, ViTP achieves good performance, posting top results on COCO object detection (AP$_{box}$ 53.9) and ADE20K semantic segmentation (mIoU 55.8). It significantly outperforms leading methods from all major pretraining paradigms, including supervised, contrastive learning, and Masked Image Modeling. On COCO instance segmentation, ViTP remains highly competitive (AP$_{mask}$ 46.7). This consistent superiority highlights that the features learned via our top-down approach are highly effective for downstream adaptation.

To more directly assess the intrinsic quality of the learned visual representations, we performed semantic segmentation on ADE20K with the frozen Vision Transformer backbone.  As detailed in Table~\ref{ade20k_freeze}, ViTP (L/16) achieves a 51.9 mIoU, surpassing other strong multimodal foundation models like AIMv2~\cite{aimv2} (51.5) and various SigLIP-based architectures. 

ViTP's leading performance in both end-to-end finetuning and frozen-feature evaluation settings provides compelling evidence for the efficacy of our top-down pretraining paradigm. This demonstrates that the VLM's understanding-based objective implicitly trains the Vision Transformer to learn more powerful and generalizable representations, confirming that top-down understanding significantly benefits bottom-up perception.

\begin{table}[!t]
\caption{Semantic segmentation performance (mIoU) on ADE20K using an UperNet~\cite{UNet} architecture with a frozen ViT backbone.}
\renewcommand\arraystretch{1.1}  
\setlength{\tabcolsep}{6pt}
\centering \footnotesize
\begin{tabular}{ccc|c}
\multirow{2}{*}{~~~Model} & \multirow{2}{*}{ViT} & \multirow{2}{*}{Size} & \multirow{2}{*}{\begin{tabular}[c]{@{}c@{}}ADE20K\\ (mIoU)\end{tabular}} \\
 &  &  &  \\ \Xhline{1pt} 
SigLIP~\cite{siglip} &  So/14 & 400M &35.2 \\ 
SigLIP2~\cite{dvqa} &  So/16 & 400M & 35.3 \\ 
LLaVA-SigLIP~\cite{llavasiglip}  & So/14 &  400M & 39.9 \\ 
COMP-SigLIP~\cite{comp} & So/14 & 400M  & 49.5 \\ 
AIMv2~\cite{aimv2}& L/14 & 300M & 51.5 \\ 
COMP-AIMv2~\cite{comp} & L/14 & 300M & 51.0 \\ \hline
ViTP & L/16 & 300M & \textbf{51.9}
\end{tabular}
\label{ade20k_freeze}
\end{table}

\begin{table}[!t]
\caption{Data efficiency on the RSAR Benchmark. ViTP demonstrates significantly stronger performance in low-data regimes compared to MIM and contrastive methods.}
\label{tab:data_efficiency}
\renewcommand\arraystretch{1.45}  
\setlength{\tabcolsep}{5pt}
  \centering \footnotesize
\begin{tabular}{l|c|cccc}
Model & 100\% & 20\% & 10\% & 5\% & 2\% \\ \Xhline{1pt}
RemoteCLIP~\cite{remoteclip} & 69.18 & \begin{tabular}[c]{@{}l@{}}\vspace{-8pt}63.14\\ \color[HTML]{5B5B5B}$_{\downarrow6.04}$\end{tabular} & \begin{tabular}[c]{@{}l@{}}\vspace{-8pt}~57.10\\ \color[HTML]{5B5B5B}$_{\downarrow12.08}$\end{tabular} & \begin{tabular}[c]{@{}l@{}}\vspace{-8pt}~47.25\\ \color[HTML]{5B5B5B}$_{\downarrow21.93}$\end{tabular} & \begin{tabular}[c]{@{}l@{}}\vspace{-8pt}~34.78\\\color[HTML]{5B5B5B}$_{\downarrow34.40}$ \end{tabular} \\
SatMAE~\cite{satmae} & 67.99 & \begin{tabular}[c]{@{}l@{}}\vspace{-8pt}61.36\\ \color[HTML]{5B5B5B}$_{\downarrow6.63}$\end{tabular} & \begin{tabular}[c]{@{}l@{}}\vspace{-8pt}~55.24\\ \color[HTML]{5B5B5B}$_{\downarrow12.75}$\end{tabular} & \begin{tabular}[c]{@{}l@{}}\vspace{-8pt}~50.76\\ \color[HTML]{5B5B5B}$_{\downarrow17.23}$\end{tabular} & \begin{tabular}[c]{@{}l@{}}\vspace{-8pt}~37.90\\ \color[HTML]{5B5B5B}$_{\downarrow30.09}$\end{tabular} \\ \hline 
\textbf{ViTP} & \textbf{72.31} &  \begin{tabular}[c]{@{}l@{}} \vspace{-8pt}\textbf{67.07}\\ \color[HTML]{5B5B5B}$_{\downarrow\textbf{5.24}}$\end{tabular} &\begin{tabular}[c]{@{}l@{}}\vspace{-8pt}~\textbf{61.68}\\ \color[HTML]{5B5B5B}$_{\downarrow\textbf{10.63}}$ \end{tabular} & \begin{tabular}[c]{@{}l@{}}\vspace{-8pt}~\textbf{56.42}\\ \color[HTML]{5B5B5B}$_{\downarrow\textbf{15.89}}$\end{tabular} & \begin{tabular}[c]{@{}l@{}}\vspace{-8pt}~\textbf{46.98}\\ \color[HTML]{5B5B5B}$_{\downarrow\textbf{25.33}}$\end{tabular} \\
\end{tabular}
\end{table}

\begin{table*}[!ht]
\caption{Model robustness to common image corruptions. ViTP demonstrates superior resilience compared to MIM and CL baselines on the REOBench benchmark (DIOR-R, mAP). It maintains the highest average performance across all corruptions and suffers the smallest performance drop ($\Delta_{\text{TP}}$) from the clean baseline. The inclusion of Visual Robustness Learning (VRL) further enhances this robustness.}
\renewcommand\arraystretch{1.2}  
\setlength{\tabcolsep}{3pt}
  \centering \footnotesize
\begin{tabular}{@{}ll|c|cccccccccccc|c|cc@{}} 
& \multirow{2}{*}{~~~Model}  & \multirow{2}{*}{Clean} & Bright & \multirow{2}{*}{Cloud} & Comp. & Data & Gauss & Gauss & \multirow{2}{*}{Haze} & Motion & \multirow{2}{*}{Rotate} & Salt  & \multirow{2}{*}{Scale} & \multirow{2}{*}{Trans} & \multirow{2}{*}{\textbf{Avg}} & \multirow{2}{*}{\textbf{$\Delta_{\text{TP}}\downarrow$} } \\
& &  & Contrast &  & Artifact & Gaps & Blur & Noise &  & Blur &  & Pepper &  &  &  & \\
\Xhline{1pt}
\multirow{4}{*}{MIM} & SatMAE~\cite{satmae}   & 62.30 & 56.84 & 57.86 & 55.80 & 58.36 & 55.38 & 58.44 & 59.34 & 56.92 & 56.60 & 53.76 & 51.58 & 60.90 & 56.82 & 5.49  \\
& ScaleMAE~\cite{scalemae}   & 70.20 & 64.80 & 65.98 & 62.50 & 64.46 & 62.58 & {63.82} & 66.10 & {63.08} & 63.44 & {60.50} & 53.08 & 68.26 & 63.22 & 6.98  \\
& RVSA~\cite{rvsa}  & {70.96} & 60.59 & 65.02 & 61.58 & 64.60 & 62.35 & 62.87 & 63.98 & 62.88 & {64.04} & 56.61  & 55.97 & {69.69} & 62.51 & 8.45 \\
& SatMAE++~\cite{satmae++} & 65.20 & 59.44 & 61.02 & 60.30 & 59.88 & 59.66 & 61.06 & 61.72 & 59.56 & 59.14 & 58.64 &48.48 & 64.70 & 59.47 & 5.73 \\
\hline
\multirow{2}{*}{CL} & RemoteCLIP~\cite{remoteclip}  & 70.20 & {66.52} & {66.62} & 63.84 & {65.40} & {63.62} & 63.68 & {66.76} & 62.66 & 63.52 & 59.16 &57.42  & 68.64 & 63.99 &  6.21 \\
& GeoRSCLIP~\cite{rs5m} & 69.80 & 66.12 & 65.34 & {65.34} & 64.96 & {63.62} & 62.90 & 66.04 & 62.02 & 62.68 & 56.04 & 57.40 & 68.10 & 63.38 & 6.42 \\ \hline
& ViTP w/o VRL & \underline{71.91} & 69.11 & 67.25 & 66.20 & 66.87 & 67.67 & 67.47 & 67.12  & 65.94 & 65.54 & 66.87 & 65.23 & 70.34 & \underline{67.13} & \underline{4.78} \\
& \textbf{ViTP} & \textbf{73.37} & \textbf{70.56} & \textbf{69.54} & \textbf{67.74} & \textbf{70.58 }& \textbf{69.05} & \textbf{68.97} & \textbf{70.85} & \textbf{67.86} & \textbf{67.10} & \textbf{67.23}  & \textbf{66.64} & \textbf{71.87} & \textbf{69.00} & \textbf{4.37}
\end{tabular}
\label{tab_det}
\end{table*}

\subsection{Data Efficiency}
 
We investigate ViTP's performance in low-data regimes by finetuning it on subsets of the RSAR training dataset, ranging from 2\% to 20\%. To fully assess the modeling potential of each method, we extend the finetuning schedule to 36 epochs for this specific study.
As shown in Table~\ref{tab:data_efficiency}, ViTP consistently outperforms SatMAE\cite{satmae} (an MIM method) and RemoteCLIP~\cite{remoteclip} (a contrastive method) across all data fractions. ViTP's advantage becomes more pronounced as the amount of training data decreases. For instance, with only 2\% of the training data, ViTP achieves 46.98 mAP, substantially outperforming SatMAE (37.90 mAP) and RemoteCLIP (34.78 mAP). Notably, ViTP finetuned on just 20\% of the data (67.07 mAP) achieves performance comparable to or exceeding many state-of-the-art methods trained on the full dataset (as shown in main paper Table~2). This superior data efficiency suggests that the rich semantic priors acquired during instruction tuning enable the model to generalize more effectively from a limited number of examples.

\subsection{Model Robustness}

While most research datasets contain curated clean imagery, real-world remote sensing images are often degraded by atmospheric conditions (e.g., clouds, haze), sensor noise, or processing artifacts. To evaluate ViTP's resilience to such degradations, we assess its robustness against 12 common image corruptions from the REOBench~\cite{reobench} benchmark on the DIOR-R dataset. We retrained our ViTP model on the DIOR-R dataset using an image size of 800$\times$800 pixels and without any test-time augmentation, matching the settings of the REOBench baseline models.
As shown in Table~\ref{tab_det}, the standard ViTP model is significantly more robust than all Masked Image Modeling (MIM) and Contrastive Learning (CL) models. It achieves the highest average mAP of 69.00 across all corruptions and the lowest performance drop ($\Delta_{\text{TP}}$) of 4.37 from its clean-image performance. We also ablate the effect of our proposed Visual Robustness Learning (VRL). The model trained without VRL already demonstrates superior robustness over the baselines. The inclusion of VRL further boosts the average performance from 67.13 to 69.00 mAP and reduces the performance degradation ($\Delta_{\text{TP}}$) from 4.78 to 4.37. This confirms that our instruction tuning paradigm, enhanced with VRL, is a highly effective technique for learning robust and comprehensive feature representations.

\begin{figure*}[t]
  \centering
  \includegraphics[width=0.95\linewidth]{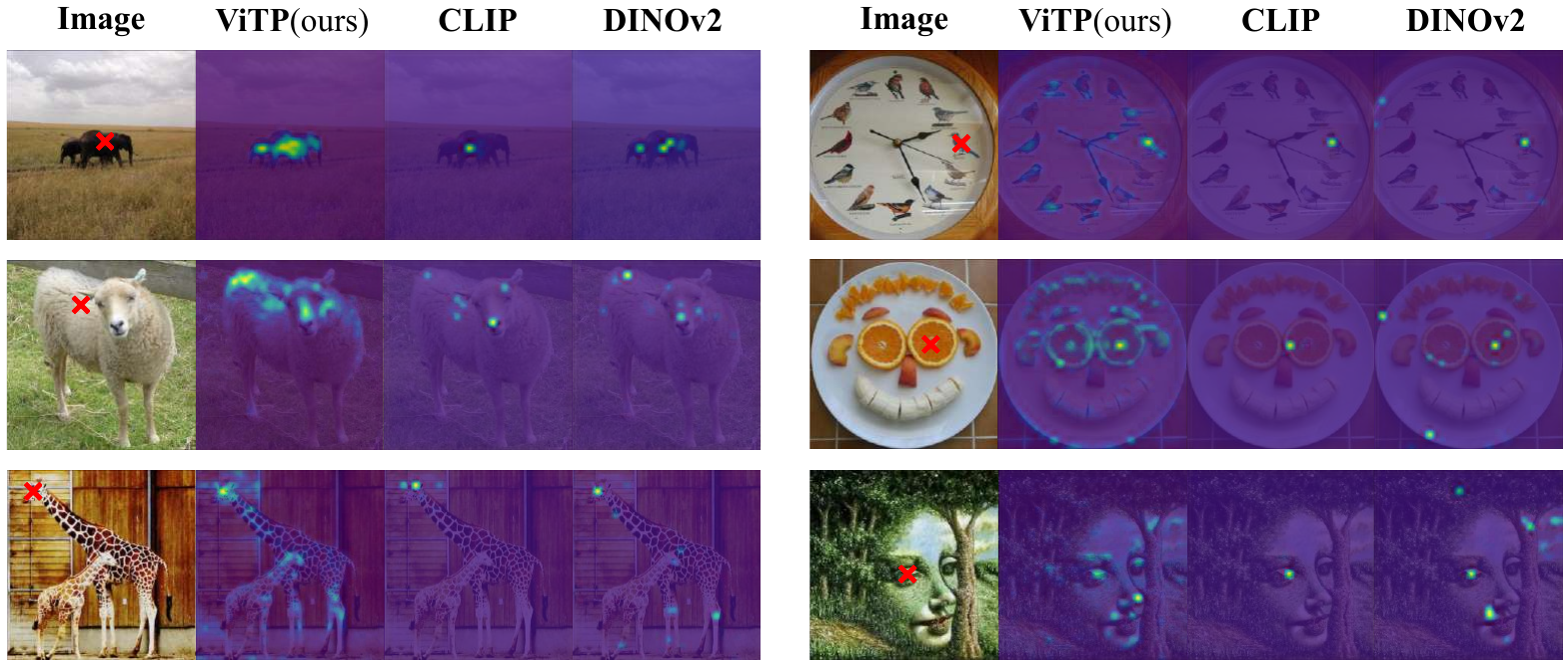}
  \caption{Comparison of self-attention activation maps from the third ViT layer for a given query patch (red cross). The models were pretrained using ViTP (ours), CLIP, and DINOv2. Left: General nature images. Right: Abstract concept images that require a higher degree of human-like understanding.}
  \label{fig:vis_1}
\end{figure*}

\begin{figure*}[t]
  \centering
  \includegraphics[width=0.95\linewidth]{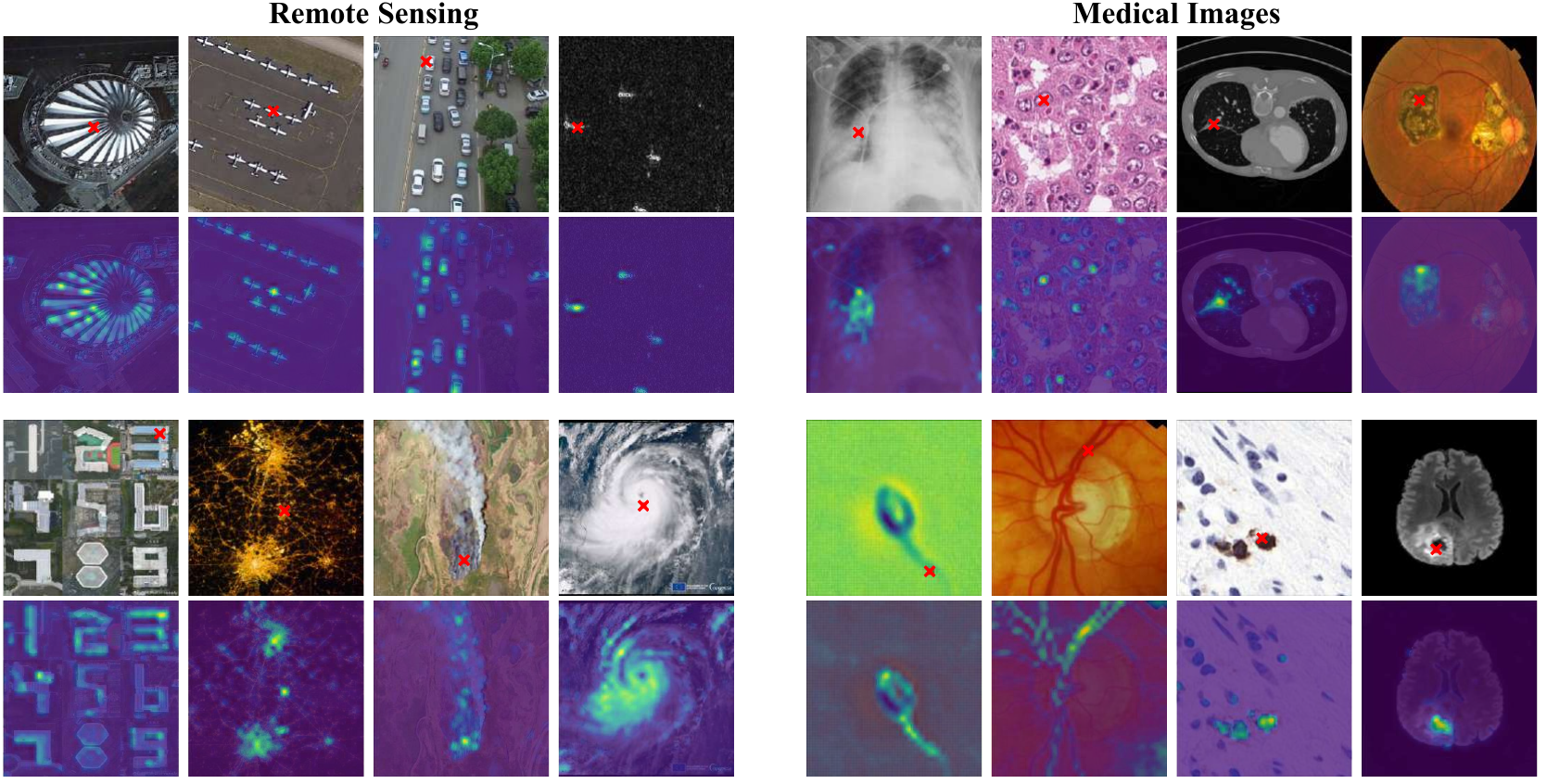}
  \caption{Self-attention activation maps for a given query patch (red cross) from the ViTP pretrained Vision Transformer. Left: ViT is pretrained with remote sensing domain data. Right: ViT is pretrained with medical domain data.}
  \label{fig:vis_2}
\end{figure*}

\section{Visualization of Self-Attention Maps}
To provide a qualitative analysis of the features learned by our ViTP backbone, we visualize its internal self-attention mechanisms. This comparison assesses how a ViT backbone, when pretrained with our Visual Instruction Pretraining (ViTP) , interprets semantic relationships within an image compared to backbones pretrained with leading visual foundation model pretraining methods: CLIP~\cite{clip} and DINOv2~\cite{dinov2}.

As shown in Figure~\ref{fig:vis_1} and Figure~\ref{fig:vis_2}, we compute the self-attention activation maps for a specific query patch (marked with a red cross). These maps show which other image regions the model ``attends to" when processing the information from that single query patch.
In our analysis, we observed the "artifacts phenomenon" (as discussed in previous works~\cite{darcet2023vision,jiang2025vision}), where some layers produce diffuse attention. We found that the third layer of the Vision Transformer provides the clearest and most ideal representation of interpretable attention across all three methods, so we visualize the outputs exclusively from this layer.

In examples like the elephants, sheep, and giraffes, the query patch (red cross) prompts the ViTP backbone to attend to other, non-local parts of the image that are semantically related. This capability is even more pronounced in abstract images where the "object" is a concept. 

We also provide more visualization on domain specific images like remote sensing and medical images in Figure~\ref{fig:vis_2}.
This analysis confirms that our ViTP successfully imbue the ViT backbone with a high-level semantic understanding, enabling it to group visual information based on conceptual meaning rather than just low-level features.

\section{Limitation and Future Work}

Despite the strong performance and efficiency of ViTP, we acknowledge several limitations that open promising avenues for future research.
A primary limitation is that ViTP's success is intrinsically linked to the quality and diversity of the instruction-following dataset. As our experiments suggest, a carefully curated data recipe is important for achieving optimal performance. The current process of collecting, filtering, and balancing datasets for specialized domains like remote sensing or medical imaging requires considerable manual effort and deep domain expertise. Future work could focus on developing more automated or semi-automated strategies for generating high-quality, domain-specific instruction data. Leveraging large language models to synthesize diverse question-answer or description-grounding pairs from unlabeled images could be a promising direction to enhance the scalability and accessibility of our pretraining paradigm.
Moreover, our current work focuses on pretraining 2D image-based Vision Transformers. Extending the ViTP framework to other data modalities, such as video and 3D point clouds, represents another exciting frontier. Adapting the instruction-following objective to these domains could unlock new capabilities in temporal understanding and 3D spatial understanding.

{
    \small
    \bibliographystyle{ieeenat_fullname}
    \bibliography{egbib}

@String(IJCV = {Int. J. Comput. Vis.})

@String(CVPR= {IEEE Conf. Comput. Vis. Pattern Recog.})

@String(ICCV= {Int. Conf. Comput. Vis.})

@String(ECCV= {Eur. Conf. Comput. Vis.})

@String(TIP  = {IEEE Trans. Image Process.})

@String(ACMMM= {ACM Int. Conf. Multimedia})

@String(ICLR = {Int. Conf. Learn. Represent.})

@String(AAAI = {AAAI})

@String(IJCV  = {IJCV})

@String(CVPR  = {CVPR})

@String(ICCV  = {ICCV})

@String(ECCV  = {ECCV})

@String(TIP   = {IEEE TIP})

@String(ACMMM = {ACM MM})

@String(ICLR  = {ICLR})

@article{rsar,
  title={Rsar: Restricted state angle resolver and rotated sar benchmark},
  author={Zhang, Xin and Yang, Xue and Li, Yuxuan and Yang, Jian and Cheng, Ming-Ming and Li, Xiang},
  journal={CVPR},
  year={2025}
}

@article{dior,
  title={Object detection in optical remote sensing images: A survey and a new benchmark},
  author={Li, Ke and Wan, Gang and Cheng, Gong and Meng, Liqiu and Han, Junwei},
  journal={ISPRS},
  year={2020},
}

@article{diorr,
  title={Anchor-free oriented proposal generator for object detection},
  author={Cheng, Gong and Wang, Jiabao and Li, Ke and Xie, Xingxing and Lang, Chunbo and Yao, Yanqing and Han, Junwei},
  journal={TGRS},
  year={2022}
}

@article{sardet,
  title={Sardet-100k: Towards open-source benchmark and toolkit for large-scale sar object detection},
  author={Li, Yuxuan and Li, Xiang and Li, Weijie and Hou, Qibin and Liu, Li and Cheng, Ming-Ming and Yang, Jian},
  journal={NeurIPS},
  year={2024}
}

@article{lsknetijcv,
  title={LSKNet: A Foundation Lightweight Backbone for Remote Sensing},
  author={Li, Yuxuan and Li, Xiang and Dai, Yimain and Hou, Qibin and Liu, Li and Liu, Yongxiang and Cheng, Ming-Ming and Yang, Jian},
  journal={IJCV},
  year={2024},
}

@inproceedings{gridrcnn,
  title={Grid r-cnn},
  author={Lu, Xin and Li, Buyu and Yue, Yuxin and Li, Quanquan and Yan, Junjie},
  booktitle={CVPR},
  year={2019}
}

@inproceedings{sparsercnn,
  title={Sparse r-cnn: End-to-end object detection with learnable proposals},
  author={Sun, Peize and Zhang, Rufeng and Jiang, Yi and Kong, Tao and Xu, Chenfeng and Zhan, Wei and Tomizuka, Masayoshi and Li, Lei and Yuan, Zehuan and Wang, Changhu and others},
  booktitle={CVPR},
  year={2021}
}

@inproceedings{dabdetr,
  title={{DAB}-{DETR}: Dynamic Anchor Boxes are Better Queries for {DETR}},
  author={Shilong Liu and Feng Li and Hao Zhang and Xiao Yang and Xianbiao Qi and Hang Su and Jun Zhu and Lei Zhang},
  booktitle={ICLR},
  year={2022},
}

@inproceedings{lsknet,
  title={Large Selective Kernel Network for Remote Sensing Object Detection},
  author={Li, Yuxuan and Hou, Qibin and Zheng, Zhaohui and Cheng, Ming-Ming and Yang, Jian and Li, Xiang},
  booktitle={ICCV},
  year={2023}
}

@inproceedings{deformabledetr,
  title={Deformable DETR: Deformable Transformers for End-to-End Object Detection},
  author={Zhu, Xizhou and Su, Weijie and Lu, Lewei and Li, Bin and Wang, Xiaogang and Dai, Jifeng},
  booktitle={ICLR},
  year={2021}
}

@inproceedings{fcos,
  title={Fcos: Fully convolutional one-stage object detection},
  author={Tian, Zhi and Shen, Chunhua and Chen, Hao and He, Tong},
  booktitle={ICCV},
  year={2019}
}

@inproceedings{detr,
  title={End-to-end object detection with transformers},
  author={Carion, Nicolas and Massa, Francisco and Synnaeve, Gabriel and Usunier, Nicolas and Kirillov, Alexander and Zagoruyko, Sergey},
  booktitle={ECCV},
  year={2020},
}

@article{gfl,
  title={Generalized focal loss: Towards efficient representation learning for dense object detection},
  author={Li, Xiang and Lv, Chengqi and Wang, Wenhai and Li, Gang and Yang, Lingfeng and Yang, Jian},
  journal={TPAMI},
  year={2022},
}

@article{zeng2024ars,
  title={ARS-DETR: Aspect Ratio-Sensitive Detection Transformer for Aerial Oriented Object Detection},
  author={Zeng, Ying and Chen, Yushi and Yang, Xue and Li, Qingyun and Yan, Junchi},
  journal={TGRS},
  year={2024},
}

@inproceedings{roitrans,
  title={Learning RoI transformer for oriented object detection in aerial images},
  author={Ding, Jian and Xue, Nan and Long, Yang and Xia, Gui-Song and Lu, Qikai},
  booktitle={CVPR},
  year={2019}
}

@inproceedings{orcnn,
  title={Oriented R-CNN for object detection},
  author={Xie, Xingxing and Cheng, Gong and Wang, Jiabao and Yao, Xiwen and Han, Junwei},
  booktitle={ICCV},
  year={2021}
}

@article{s2anet,
  title={Align deep features for oriented object detection},
  author={Han, Jiaming and Ding, Jian and Li, Jie and Xia, Gui-Song},
  journal={TGRS},
  year={2021},
}

@inproceedings{redet,
  title={Redet: A rotation-equivariant detector for aerial object detection},
  author={Han, Jiaming and Ding, Jian and Xue, Nan and Xia, Gui-Song},
  booktitle={CVPR},
  year={2021}
}

@inproceedings{r3det,
  title={R3det: Refined single-stage detector with feature refinement for rotating object},
  author={Yang, Xue and Yan, Junchi and Feng, Ziming and He, Tao},
  booktitle={AAAI},
  year={2021}
}

@inproceedings{retina,
  title={Focal loss for dense object detection},
  author={Lin, Tsung-Yi and Goyal, Priya and Girshick, Ross and He, Kaiming},
  booktitle={ICCV},
  year={2017}
}

@article{S2Looking,
  title={S2Looking: A satellite side-looking dataset for building change detection},
  author={Shen, Li and Lu, Yao and Chen, Hao and Wei, Hao and Xie, Donghai and Yue, Jiabao and Chen, Rui and Lv, Shouye and Jiang, Bitao},
  journal={Remote Sensing},
  year={2021} 
}

@article{llava,
  title={Visual instruction tuning},
  author={Liu, Haotian and Li, Chunyuan and Wu, Qingyang and Lee, Yong Jae},
  journal={NeurIPS}, 
  year={2024}
}

@article{skysense,
  title={Skysense: A multi-modal remote sensing foundation model towards universal interpretation for earth observation imagery},
  author={Guo, Xin and Lao, Jiangwei and Dang, Bo and Zhang, Yingying and Yu, Lei and Ru, Lixiang and Zhong, Liheng and Huang, Ziyuan and Wu, Kang and Hu, Dingxiang and others},
  journal={arXiv},
  year={2023}
}

@article{RinMo,
  author={Sun, Xian and Wang, Peijin and Lu, Wanxuan and Zhu, Zicong and Lu, Xiaonan and He, Qibin and Li, Junxi and Rong, Xuee and Yang, Zhujun and Chang, Hao and He, Qinglin and Yang, Guang and Wang, Ruiping and Lu, Jiwen and Fu, Kun},
  journal={TGRS}, 
  title={RingMo: A Remote Sensing Foundation Model With Masked Image Modeling}, 
  year={2023}}

@article{CGNet,
  title={Change guiding network: Incorporating change prior to guide change detection in remote sensing imagery},
  author={Han, Chengxi and Wu, Chen and Guo, Haonan and Hu, Meiqi and Li, Jiepan and Chen, Hongruixuan},
  journal={IEEE Journal of Selected Topics in Applied Earth Observations and Remote Sensing},
  year={2023},
}

@inproceedings{changeformer,
  title={A transformer-based siamese network for change detection},
  author={Bandara, Wele Gedara Chaminda and Patel, Vishal M},
  booktitle={IEEE International Geoscience and Remote Sensing Symposium}, 
  year={2022}, 
}

@article{changer,
  title={Changer: Feature interaction is what you need for change detection},
  author={Fang, Sheng and Li, Kaiyu and Li, Zhe},
  journal={TGRS},
  year={2023} 
}

@article{snunet,
  title={SNUNet-CD: A densely connected Siamese network for change detection of VHR images},
  author={Fang, Sheng and Li, Kaiyu and Shao, Jinyuan and Li, Zhe},
  journal={IEEE Geoscience and Remote Sensing Letters},
  year={2021}
}

@article{bit,
  title={Remote sensing image change detection with transformers},
  author={Chen, Hao and Qi, Zipeng and Shi, Zhenwei},
  journal={TGRS},
  year={2021}
}

@article{lin2024diformer,
  title={DiFormer: A Difference Transformer Network for Remote Sensing Change Detection},
  author={Lin, Hui and Hang, Renlong and Wang, Shanmin and Liu, Qingshan},
  journal={IEEE Geoscience and Remote Sensing Letters},
  year={2024}
}

@ARTICLE{RSSFormer,
  author={Xu, Rongtao and Wang, Changwei and Zhang, Jiguang and Xu, Shibiao and Meng, Weiliang and Zhang, Xiaopeng},
  journal={TIP}, 
  title={RSSFormer: Foreground Saliency Enhancement for Remote Sensing Land-Cover Segmentation}, 
  year={2023},}

@inproceedings{SegFormer,
  title={{SegFormer}: Simple and Efficient Design for Semantic Segmentation with Transformers},
  author={Xie, Enze and Wang, Wenhai and Yu, Zhiding and Anandkumar, Anima and Alvarez, Jose M and Luo, Ping},
  booktitle={NeurIPS},
  year={2021}
}

@inproceedings{UNet,
  title={Unified Perceptual Parsing for Scene Understanding},
  author={Xiao, Tete and Liu, Yingcheng and Zhou, Bolei and Jiang, Yuning and Sun, Jian},
  booktitle={ECCV},
  year={2018},
}

@inproceedings{PSPNet,
  title={Pyramid scene parsing network},
  author={Zhao, Hengshuang and Shi, Jianping and Qi, Xiaojuan and Wang, Xiaogang and Jia, Jiaya},
  booktitle={CVPR},
  year={2017}
}

@article{ABCNet,
    title = {{ABCNet}: Attentive bilateral contextual network for efficient semantic segmentation of Fine-Resolution remotely sensed imagery},
    journal = {ISPRS},
    year = {2021},
    author = {Rui Li and Shunyi Zheng and Ce Zhang and Chenxi Duan and Libo Wang and Peter M. Atkinson},
}

@inproceedings{Segmenter,
  title={Segmenter: Transformer for semantic segmentation},
  author={Strudel, Robin and Garcia, Ricardo and Laptev, Ivan and Schmid, Cordelia},
  booktitle={ICCV},
  year={2021}
}

@article{UNetFormer,
    title = {{UNetFormer}: A {UNet}-like transformer for efficient semantic segmentation of remote sensing urban scene imagery},
    journal = {ISPRS},
    year = {2022},
    author = {Libo Wang and Rui Li and Ce Zhang and Shenghui Fang and Chenxi Duan and Xiaoliang Meng and Peter M. Atkinson},
}

@inproceedings{Swin,
  title={Swin transformer: Hierarchical vision transformer using shifted windows},
  author={Liu, Ze and Lin, Yutong and Cao, Yue and Hu, Han and Wei, Yixuan and Zhang, Zheng and Lin, Stephen and Guo, Baining},
  booktitle={CVPR},
  year={2021}
}

@inproceedings{cascade_rcnn,  
    author={Cai, Zhaowei and Vasconcelos, Nuno},  
    booktitle={CVPR},   
    title={Cascade {R-CNN}: Delving Into High Quality Object Detection},   
    year={2018}}

@inproceedings{dota,  
    author={Xia, Gui-Song and Bai, Xiang and Ding, Jian and Zhu, Zhen and Belongie, Serge and Luo, Jiebo and Datcu, Mihai and Pelillo, Marcello and Zhang, Liangpei},  
    booktitle={CVPR},   
    title={{DOTA}: A Large-Scale Dataset for Object Detection in Aerial Images},   
    year={2018},  }

@article{qwen,
  title={Qwen technical report},
  author={Bai, Jinze and Bai, Shuai and Chu, Yunfei and Cui, Zeyu and Dang, Kai and Deng, Xiaodong and Fan, Yang and Ge, Wenbin and Han, Yu and Huang, Fei and others},
  journal={arXiv},
  year={2023}
}

@article{qwen25vl,
  title={Qwen2. 5-vl technical report},
  author={Bai, Shuai and Chen, Keqin and Liu, Xuejing and Wang, Jialin and Ge, Wenbin and Song, Sibo and Dang, Kai and Wang, Peng and Wang, Shijie and Tang, Jun and others},
  journal={arXiv},
  year={2025}
}

@inproceedings{internvl,
  title={Internvl: Scaling up vision foundation models and aligning for generic visual-linguistic tasks},
  author={Chen, Zhe and Wu, Jiannan and Wang, Wenhai and Su, Weijie and Chen, Guo and Xing, Sen and Zhong, Muyan and Zhang, Qinglong and Zhu, Xizhou and Lu, Lewei and others},
  booktitle={CVPR}, 
  year={2024}
}

@article{gemini,
  title={Gemini: a family of highly capable multimodal models},
  author={Team, Gemini and Anil, Rohan and Borgeaud, Sebastian and Alayrac, Jean-Baptiste and Yu, Jiahui and Soricut, Radu and Schalkwyk, Johan and Dai, Andrew M and Hauth, Anja and Millican, Katie and others},
  journal={arXiv},
  year={2023}
}

@inproceedings{mae,
  title={Masked autoencoders are scalable vision learners},
  author={He, Kaiming and Chen, Xinlei and Xie, Saining and Li, Yanghao and Doll{\'a}r, Piotr and Girshick, Ross},
  booktitle={CVPR},
  year={2022}
}

@article{beit,
  title={Beit: Bert pre-training of image transformers},
  author={Bao, Hangbo and Dong, Li and Piao, Songhao and Wei, Furu},
  journal={arXiv},
  year={2021}
}

@inproceedings{moco,
  title={Momentum contrast for unsupervised visual representation learning},
  author={He, Kaiming and Fan, Haoqi and Wu, Yuxin and Xie, Saining and Girshick, Ross},
  booktitle={CVPR},
  year={2020}
}

@article{dinov2,
  title={Dinov2: Learning robust visual features without supervision},
  author={Oquab, Maxime and Darcet, Timoth{\'e}e and Moutakanni, Th{\'e}o and Vo, Huy and Szafraniec, Marc and Khalidov, Vasil and Fernandez, Pierre and Haziza, Daniel and Massa, Francisco and El-Nouby, Alaaeldin and others},
  journal={arXiv},
  year={2023}
}

@ARTICLE{RVSA,
  author={Wang, Di and Zhang, Qiming and Xu, Yufei and Zhang, Jing and Du, Bo and Tao, Dacheng and Zhang, Liangpei},
  journal={TGRS}, 
  title={Advancing Plain Vision Transformer Toward Remote Sensing Foundation Model}, 
  year={2023}}

@inproceedings{clip,
  title={Learning transferable visual models from natural language supervision},
  author={Radford, Alec and Kim, Jong Wook and Hallacy, Chris and Ramesh, Aditya and Goh, Gabriel and Agarwal, Sandhini and Sastry, Girish and Askell, Amanda and Mishkin, Pamela and Clark, Jack and others},
  booktitle={ICML},
  year={2021}
}

@article{vit,
  title={An image is worth 16x16 words: Transformers for image recognition at scale},
  author={Dosovitskiy, Alexey and Beyer, Lucas and Kolesnikov, Alexander and Weissenborn, Dirk and Zhai, Xiaohua and Unterthiner, Thomas and Dehghani, Mostafa and Minderer, Matthias and Heigold, Georg and Gelly, Sylvain and others},
  journal={arXiv},
  year={2020}
}

@article{byol,
  title={Bootstrap your own latent-a new approach to self-supervised learning},
  author={Grill, Jean-Bastien and Strub, Florian and Altch{\'e}, Florent and Tallec, Corentin and Richemond, Pierre and Buchatskaya, Elena and Doersch, Carl and Avila Pires, Bernardo and Guo, Zhaohan and Gheshlaghi Azar, Mohammad and others},
  journal={NeurIPS},
  year={2020}
}

@article{satmae,
  title={Satmae: Pre-training transformers for temporal and multi-spectral satellite imagery},
  author={Cong, Yezhen and Khanna, Samar and Meng, Chenlin and Liu, Patrick and Rozi, Erik and He, Yutong and Burke, Marshall and Lobell, David and Ermon, Stefano},
  journal={NeurIPS},
  year={2022}
}

@inproceedings{scalemae,
  title={Scale-mae: A scale-aware masked autoencoder for multiscale geospatial representation learning},
  author={Reed, Colorado J and Gupta, Ritwik and Li, Shufan and Brockman, Sarah and Funk, Christopher and Clipp, Brian and Keutzer, Kurt and Candido, Salvatore and Uyttendaele, Matt and Darrell, Trevor},
  booktitle={ICCV},
  year={2023}
}

@inproceedings{seco,
  title={Seco: Exploring sequence supervision for unsupervised representation learning},
  author={Yao, Ting and Zhang, Yiheng and Qiu, Zhaofan and Pan, Yingwei and Mei, Tao},
  booktitle={AAAI},
  year={2021}
}

@inproceedings{caco,
  title={Change-aware sampling and contrastive learning for satellite images},
  author={Mall, Utkarsh and Hariharan, Bharath and Bala, Kavita},
  booktitle={CVPR},
  year={2023}
}

@article{cmid,
  title={Cmid: A unified self-supervised learning framework for remote sensing image understanding},
  author={Muhtar, Dilxat and Zhang, Xueliang and Xiao, Pengfeng and Li, Zhenshi and Gu, Feng},
  journal={TGRS},
  year={2023}
}

@inproceedings{gfm,
  title={Towards geospatial foundation models via continual pretraining},
  author={Mendieta, Mat{\'\i}as and Han, Boran and Shi, Xingjian and Zhu, Yi and Chen, Chen},
  booktitle={ICCV},
  year={2023}
}

@article{selectivemae,
  title={Harnessing massive satellite imagery with efficient masked image modeling},
  author={Wang, F and Wang, H and Wang, D and Guo, Z and Zhong, Z and Lan, L and Yang, W and Zhang, J},
  journal={arXiv},
  year={2025}
}

@article{fasterrcnn,
  title={Faster r-cnn: Towards real-time object detection with region proposal networks},
  author={Ren, Shaoqing and He, Kaiming and Girshick, Ross and Sun, Jian},
  journal={NeurIPS},
  year={2015}
}

@article{billionfm,
  title={A billion-scale foundation model for remote sensing images},
  author={Cha, Keumgang and Seo, Junghoon and Lee, Taekyung},
  journal={arXiv},
  year={2023}
}

@inproceedings{dcfl,
  title={Dynamic coarse-to-fine learning for oriented tiny object detection},
  author={Xu, Chang and Ding, Jian and Wang, Jinwang and Yang, Wen and Yu, Huai and Yu, Lei and Xia, Gui-Song},
  booktitle={CVPR},
  year={2023}
}

@inproceedings{satlas,
  title={Satlaspretrain: A large-scale dataset for remote sensing image understanding},
  author={Bastani, Favyen and Wolters, Piper and Gupta, Ritwik and Ferdinando, Joe and Kembhavi, Aniruddha},
  booktitle={ICCV},
  year={2023}
}

@article{gghf,
  title={A general Gaussian heatmap label assignment for arbitrary-oriented object detection},
  author={Huang, Zhanchao and Li, Wei and Xia, Xiang-Gen and Tao, Ran},
  journal={TIP},
  year={2022}
}

@article{tov,
  title={TOV: The original vision model for optical remote sensing image understanding via self-supervised learning},
  author={Tao, Chao and Qi, Ji and Zhang, Guo and Zhu, Qing and Lu, Weipeng and Li, Haifeng},
  journal={IEEE Journal of Selected Topics in Applied Earth Observations and Remote Sensing},
  year={2023}
}

@inproceedings{gassl,
  title={Geography-aware self-supervised learning},
  author={Ayush, Kumar and Uzkent, Burak and Meng, Chenlin and Tanmay, Kumar and Burke, Marshall and Lobell, David and Ermon, Stefano},
  booktitle={ICCV},
  year={2021}
}

@article{ssl4eo,
  title={SSL4EO-S12: A large-scale multimodal, multitemporal dataset for self-supervised learning in Earth observation [Software and Data Sets]},
  author={Wang, Yi and Braham, Nassim Ait Ali and Xiong, Zhitong and Liu, Chenying and Albrecht, Conrad M and Zhu, Xiao Xiang},
  journal={IEEE Geoscience and Remote Sensing Magazine},
  year={2023},
  publisher={IEEE}
}

@inproceedings{ma3e,
  title={Masked angle-aware autoencoder for remote sensing images},
  author={Li, Zhihao and Hou, Biao and Ma, Siteng and Wu, Zitong and Guo, Xianpeng and Ren, Bo and Jiao, Licheng},
  booktitle={ECCV},
  year={2024},
  organization={Springer}
}

@inproceedings{atss,
  title={Bridging the gap between anchor-based and anchor-free detection via adaptive training sample selection},
  author={Zhang, Shifeng and Chi, Cheng and Yao, Yongqiang and Lei, Zhen and Li, Stan Z},
  booktitle={CVPR},
  year={2020}
}

@article{kld,
  title={Learning high-precision bounding box for rotated object detection via kullback-leibler divergence},
  author={Yang, Xue and Yang, Xiaojiang and Yang, Jirui and Ming, Qi and Wang, Wentao and Tian, Qi and Yan, Junchi},
  journal={NeurIPS},
  year={2021}
}

@inproceedings{reppoints,
  title={Oriented reppoints for aerial object detection},
  author={Li, Wentong and Chen, Yijie and Hu, Kaixuan and Zhu, Jianke},
  booktitle={CVPR},
  year={2022}
}

@inproceedings{sasm,
  title={Shape-adaptive selection and measurement for oriented object detection},
  author={Hou, Liping and Lu, Ke and Xue, Jian and Li, Yuqiu},
  booktitle={AAAI},
  year={2022}
}

@article{sm3det,
  title={Sm3det: A unified model for multi-modal remote sensing object detection},
  author={Li, Yuxuan and Li, Xiang and Li, Yunheng and Zhang, Yicheng and Dai, Yimian and Hou, Qibin and Cheng, Ming-Ming and Yang, Jian},
  journal={arXiv},
  year={2024}
}

@article{gpt4,
  title={Gpt-4 technical report},
  author={Achiam, Josh and Adler, Steven and Agarwal, Sandhini and Ahmad, Lama and Akkaya, Ilge and Aleman, Florencia Leoni and Almeida, Diogo and Altenschmidt, Janko and Altman, Sam and Anadkat, Shyamal and others},
  journal={arXiv},
  year={2023}
}

@inproceedings{geochat,
  title={Geochat: Grounded large vision-language model for remote sensing},
  author={Kuckreja, Kartik and Danish, Muhammad Sohail and Naseer, Muzammal and Das, Abhijit and Khan, Salman and Khan, Fahad Shahbaz},
  booktitle={CVPR},
  year={2024}
}

@inproceedings{vhm,
  title={Vhm: Versatile and honest vision language model for remote sensing image analysis},
  author={Pang, Chao and Weng, Xingxing and Wu, Jiang and Li, Jiayu and Liu, Yi and Sun, Jiaxing and Li, Weijia and Wang, Shuai and Feng, Litong and Xia, Gui-Song and others},
  booktitle={AAAI},
  year={2025}
}

@inproceedings{lin2017feature,
  title={Feature pyramid networks for object detection},
  author={Lin, Tsung-Yi and Doll{\'a}r, Piotr and Girshick, Ross and He, Kaiming and Hariharan, Bharath and Belongie, Serge},
  booktitle={CVPR},
  year={2017}
}

@article{rsvqa,
  title={RSVQA: Visual question answering for remote sensing data},
  author={Lobry, Sylvain and Marcos, Diego and Murray, Jesse and Tuia, Devis},
  journal={TGRS},
  year={2020},
}

@article{skysensegpt,
  title={Skysensegpt: A fine-grained instruction tuning dataset and model for remote sensing vision-language understanding},
  author={Luo, Junwei and Pang, Zhen and Zhang, Yongjun and Wang, Tingzhu and Wang, Linlin and Dang, Bo and Lao, Jiangwei and Wang, Jian and Chen, Jingdong and Tan, Yihua and others},
  journal={arXiv},
  year={2024}
}

@article{miniinternvl,
  title={Mini-internvl: a flexible-transfer pocket multi-modal model with 5\% parameters and 90\% performance},
  author={Gao, Zhangwei and Chen, Zhe and Cui, Erfei and Ren, Yiming and Wang, Weiyun and Zhu, Jinguo and Tian, Hao and Ye, Shenglong and He, Junjun and Zhu, Xizhou and others},
  journal={Visual Intelligence},
  year={2024}
}

@article{zavras2025gaia,
  title={GAIA: A global, multi-modal, multi-scale vision-language dataset for remote sensing image analysis},
  author={Zavras, Angelos and Michail, Dimitrios and Zhu, Xiao Xiang and Demir, Beg{\"u}m and Papoutsis, Ioannis},
  journal={arXiv},
  year={2025}
}

@article{LevirCC,
  title={Remote sensing image change captioning with dual-branch transformers: A new method and a large scale dataset},
  author={Liu, Chenyang and Zhao, Rui and Chen, Hao and Zou, Zhengxia and Shi, Zhenwei},
  journal={TGRS},
  year={2022}
}

@article{millionaid,
  title={On creating benchmark dataset for aerial image interpretation: Reviews, guidances, and million-aid},
  author={Long, Yang and Xia, Gui-Song and Li, Shengyang and Yang, Wen and Yang, Michael Ying and Zhu, Xiao Xiang and Zhang, Liangpei and Li, Deren},
  journal={IEEE Journal of selected topics in applied earth observations and remote sensing},
  year={2021}
}

@misc{sentinel12_image_pairs,
  author = {Paritosh Tiwari},
  title = {Sentinel-1\&2 Image Pairs (SAR \& Optical)},
  howpublished = {sentinel12-image-pairs-segregated-by-terrain/code?datasetId=1201791},
  year={2025},
  note = {Accessed: 2025-08-22},
}

@misc{isprs_sar,
  author = {ISPRS},
  title = {Classification of Multiscale Marine Phenomenon in SAR Images},
  howpublished = {isprs2024tc1},
  year={2025},
  note = {Accessed: 2025-08-22},
}

@article{vrsbench,
  title={Vrsbench: A versatile vision-language benchmark dataset for remote sensing image understanding},
  author={Li, Xiang and Ding, Jian and Elhoseiny, Mohamed},
  journal={NeurIPS},
  year={2024}
}

@article{diorrsvg,
  title={Rsvg: Exploring data and models for visual grounding on remote sensing data},
  author={Zhan, Yang and Xiong, Zhitong and Yuan, Yuan},
  journal={TGRS},
  year={2023},
}

@inproceedings{rsvg,
  title={Visual grounding in remote sensing images},
  author={Sun, Yuxi and Feng, Shanshan and Li, Xutao and Ye, Yunming and Kang, Jian and Huang, Xu},
  booktitle={ACMMM},
  year={2022}
}

@article{loveda,
  title={LoveDA: A remote sensing land-cover dataset for domain adaptive semantic segmentation},
  author={Wang, Junjue and Zheng, Zhuo and Ma, Ailong and Lu, Xiaoyan and Zhong, Yanfei},
  journal={arXiv},
  year={2021}
}

@inproceedings{isaid,
  title={isaid: A large-scale dataset for instance segmentation in aerial images},
  author={Waqas Zamir, Syed and Arora, Aditya and Gupta, Akshita and Khan, Salman and Sun, Guolei and Shahbaz Khan, Fahad and Zhu, Fan and Shao, Ling and Xia, Gui-Song and Bai, Xiang},
  booktitle={CVPR workshops},
  year={2019}
}

@article{ssdd,
  title={SAR ship detection dataset (SSDD): Official release and comprehensive data analysis},
  author={Zhang, Tianwen and Zhang, Xiaoling and Li, Jianwei and Xu, Xiaowo and Wang, Baoyou and Zhan, Xu and Xu, Yanqin and Ke, Xiao and Zeng, Tianjiao and Su, Hao and others},
  journal={Remote Sensing},
  year={2021}
}

@inproceedings{satmae++,
  title={Rethinking transformers pre-training for multi-spectral satellite imagery},
  author={Noman, Mubashir and Naseer, Muzammal and Cholakkal, Hisham and Anwer, Rao Muhammad and Khan, Salman and Khan, Fahad Shahbaz},
  booktitle={CVPR},
  year={2024}
}

@article{remoteclip,
  title={Remoteclip: A vision language foundation model for remote sensing},
  author={Liu, Fan and Chen, Delong and Guan, Zhangqingyun and Zhou, Xiaocong and Zhu, Jiale and Ye, Qiaolin and Fu, Liyong and Zhou, Jun},
  journal={TGRS},
  year={2024}
}

@article{rs5m,
  title={Rs5m and georsclip: A large scale vision-language dataset and a large vision-language model for remote sensing},
  author={Zhang, Zilun and Zhao, Tiancheng and Guo, Yulong and Yin, Jianwei},
  journal={TGRS},
  year={2024}
}

@article{UAVid,
title = {{UAVid}: A semantic segmentation dataset for UAV imagery},
journal = {ISPRS},
year = {2020},
author = {Ye Lyu and George Vosselman and Gui-Song Xia and Alper Yilmaz and Michael Ying Yang},
}

@article{CANet,
title = {Real-time Semantic Segmentation with Context Aggregation Network},
journal = {ISPRS},
year = {2021},
author = {Michael Ying Yang and Saumya Kumaar and Ye Lyu and Francesco Nex},
}

@inproceedings{maskformer,
  title={Masked-attention mask transformer for universal image segmentation},
  author={Cheng, Bowen and Misra, Ishan and Schwing, Alexander G and Kirillov, Alexander and Girdhar, Rohit},
  booktitle={CVPR},
  year={2022}
}

@inproceedings{CoaT,
  title={Co-scale conv-attentional image transformers},
  author={Xu, Weijian and Xu, Yifan and Chang, Tyler and Tu, Zhuowen},
  booktitle={ICCV},
  year={2021}
}

@inproceedings{mpformer,
  title={Mp-former: Mask-piloted transformer for image segmentation},
  author={Zhang, Hao and Li, Feng and Xu, Huaizhe and Huang, Shijia and Liu, Shilong and Ni, Lionel M and Zhang, Lei},
  booktitle={CVPR}, 
  year={2023}
}

@ARTICLE{DecoupleNet,
  author={Lu, Wei and Chen, Si-Bao and Shu, Qing-Ling and Tang, Jin and Luo, Bin},
  journal={TGRS}, 
  title={DecoupleNet: A Lightweight Backbone Network With Efficient Feature Decoupling for Remote Sensing Visual Tasks}, 
  year={2024}, }

@inproceedings{denseclip,
  title={Denseclip: Language-guided dense prediction with context-aware prompting},
  author={Rao, Yongming and Zhao, Wenliang and Chen, Guangyi and Tang, Yansong and Zhu, Zheng and Huang, Guan and Zhou, Jie and Lu, Jiwen},
  booktitle={CVPR}, 
  year={2022}
}

@inproceedings{ocrnet,
  title={Object-contextual representations for semantic segmentation},
  author={Yuan, Yuhui and Chen, Xilin and Wang, Jingdong},
  booktitle={ECCV},
  year={2020}
}

@inproceedings{deeplabv3+,
  title={Encoder-decoder with atrous separable convolution for semantic image segmentation},
  author={Chen, Liang-Chieh and Zhu, Yukun and Papandreou, George and Schroff, Florian and Adam, Hartwig},
  booktitle={ECCV},
  year={2018}
}

@inproceedings{instaboost,
  title={Instaboost: Boosting instance segmentation via probability map guided copy-pasting},
  author={Fang, Hao-Shu and Sun, Jianhua and Wang, Runzhong and Gou, Minghao and Li, Yong-Lu and Lu, Cewu},
  booktitle={ICCV},
  year={2019}
}

@article{rsprompter,
  title={RSPrompter: Learning to prompt for remote sensing instance segmentation based on visual foundation model},
  author={Chen, Keyan and Liu, Chenyang and Chen, Hao and Zhang, Haotian and Li, Wenyuan and Zou, Zhengxia and Shi, Zhenwei},
  journal={TGRS},
  year={2024}
}

@article{hqisnet,
  title={HQ-ISNet: High-quality instance segmentation for remote sensing imagery},
  author={Su, Hao and Wei, Shunjun and Liu, Shan and Liang, Jiadian and Wang, Chen and Shi, Jun and Zhang, Xiaoling},
  journal={Remote Sensing},
  year={2020}
}

@inproceedings{mask2former,
  title={Masked-attention mask transformer for universal image segmentation},
  author={Cheng, Bowen and Misra, Ishan and Schwing, Alexander G and Kirillov, Alexander and Girdhar, Rohit},
  booktitle={CVPR},
  year={2022}
}

@inproceedings{scnet,
  title={Scnet: Training inference sample consistency for instance segmentation},
  author={Vu, Thang and Kang, Haeyong and Yoo, Chang D},
  booktitle={AAAI},
  year={2021}
}

@inproceedings{condinst,
  title={Conditional convolutions for instance segmentation},
  author={Tian, Zhi and Shen, Chunhua and Chen, Hao},
  booktitle={ECCV},
  year={2020}
}

@inproceedings{boxinst,
  title={Boxinst: High-performance instance segmentation with box annotations},
  author={Tian, Zhi and Shen, Chunhua and Wang, Xinlong and Chen, Hao},
  booktitle={CVPR},
  year={2021}
}

@inproceedings{sam,
  title={Segment anything},
  author={Kirillov, Alexander and Mintun, Eric and Ravi, Nikhila and Mao, Hanzi and Rolland, Chloe and Gustafson, Laura and Xiao, Tete and Whitehead, Spencer and Berg, Alexander C and Lo, Wan-Yen and others},
  booktitle={ICCV},
  year={2023}
}

@article{catnet,
  title={Catnet: context aggregation network for instance segmentation in remote sensing images},
  author={Liu, Ye and Li, Huifang and Hu, Chao and Luo, Shuang and Shen, Huanfeng and Chen, C},
  journal={arXiv},
  year={2021}
}

@inproceedings{maskrcnn,
  title={Mask r-cnn},
  author={He, Kaiming and Gkioxari, Georgia and Doll{\'a}r, Piotr and Girshick, Ross},
  booktitle={ICCV},
  year={2017}
}

@article{vitadapter,
  title={Vision Transformer Adapter for Dense Predictions},
  author={Chen, Zhe and Duan, Yuchen and Wang, Wenhai and He, Junjun and Lu, Tong and Dai, Jifeng and Qiao, Yu},
  journal={arXiv},
  year={2022}
}

@article{internvl25,
  title={Expanding Performance Boundaries of Open-Source Multimodal Models with Model, Data, and Test-Time Scaling},
  author={Chen, Zhe and Wang, Weiyun and Cao, Yue and Liu, Yangzhou and Gao, Zhangwei and Cui, Erfei and Zhu, Jinguo and Ye, Shenglong and Tian, Hao and Liu, Zhaoyang and others},
  journal={arXiv},
  year={2024}
}

@article{transformer,
  title={Attention is all you need},
  author={Vaswani, Ashish and Shazeer, Noam and Parmar, Niki and Uszkoreit, Jakob and Jones, Llion and Gomez, Aidan N and Kaiser, {\L}ukasz and Polosukhin, Illia},
  journal={NeurIPS},
  year={2017}
}

@article{llmrotate,
  title={A simple aerial detection baseline of multimodal language models},
  author={Li, Qingyun and Chen, Yushi and Shu, Xinya and Chen, Dong and He, Xin and Yu, Yi and Yang, Xue},
  journal={arXiv},
  year={2025}
}

@article{denodet,
  title={Denodet: Attention as deformable multi-subspace feature denoising for target detection in sar images},
  author={Dai, Yimian and Zou, Minrui and Li, Yuxuan and Li, Xiang and Ni, Kang and Yang, Jian},
  journal={IEEE Transactions on Aerospace and Electronic Systems},
  year={2024}
}

@article{denodetv2,
  title={DenoDet V2: Phase-Amplitude Cross Denoising for SAR Object Detection},
  author={Ni, Kang and Zou, Minrui and Li, Yuxuan and Li, Xiang and Guo, Kehua and Cheng, Ming-Ming and Dai, Yimian},
  journal={arXiv},
  year={2025}
}

@article{saratr,
  title={SARATR-X: Towards building a foundation model for SAR target recognition},
  author={Li, Weijie and Yang, Wei and Hou, Yuenan and Liu, Li and Liu, Yongxiang and Li, Xiang},
  journal={TIP},
  year={2025}
}

@article{reobench,
  title={REOBench: Benchmarking Robustness of Earth Observation Foundation Models},
  author={Li, Xiang and Tao, Yong and Zhang, Siyuan and Liu, Siwei and Xiong, Zhitong and Luo, Chunbo and Liu, Lu and Pechenizkiy, Mykola and Zhu, Xiao Xiang and Huang, Tianjin},
  journal={arXiv},
  year={2025}
}

@article{srcdnet,
  title={Super-resolution-based change detection network with stacked attention module for images with different resolutions},
  author={Liu, Mengxi and Shi, Qian and Marinoni, Andrea and He, Da and Liu, Xiaoping and Zhang, Liangpei},
  journal={TGRS},
  year={2021}
}

@article{clnet,
  title={CLNet: Cross-layer convolutional neural network for change detection in optical remote sensing imagery},
  author={Zheng, Zhi and Wan, Yi and Zhang, Yongjun and Xiang, Sizhe and Peng, Daifeng and Zhang, Bin},
  journal={ISPRS},
  year={2021},
}

@article{escnet,
  title={ESCNet: An end-to-end superpixel-enhanced change detection network for very-high-resolution remote sensing images},
  author={Zhang, Hongyan and Lin, Manhui and Yang, Guangyi and Zhang, Liangpei},
  journal={IEEE Transactions on Neural Networks and Learning Systems},
  year={2021}
}

@article{DSAMNet,
  title={A deeply supervised attention metric-based network and an open aerial image dataset for remote sensing change detection},
  author={Shi, Qian and Liu, Mengxi and Li, Shengchen and Liu, Xiaoping and Wang, Fei and Zhang, Liangpei},
  journal={TGRS},
  year={2021}
}

@article{gcdddpm,
  title={GCD-DDPM: A generative change detection model based on difference-feature-guided DDPM},
  author={Wen, Yihan and Ma, Xianping and Zhang, Xiaokang and Pun, Man-On},
  journal={TGRS},
  year={2024},
}

@article{cdcontrast,
  title={Change detection based on supervised contrastive learning for high-resolution remote sensing imagery},
  author={Wang, Jue and Zhong, Yanfei and Zhang, Liangpei},
  journal={TGRS},
  year={2023}
}

@inproceedings{ddpmcd,
  title={DDPM-CD: Denoising diffusion probabilistic models as feature extractors for remote sensing change detection},
  author={Bandara, Wele Gedara Chaminda and Nair, Nithin Gopalakrishnan and Patel, Vishal M},
  booktitle={IEEE/CVF Winter Conference on Applications of Computer Vision},
  year={2025},
}

@article{DMNet,
  title={ESR-DMNet: Enhanced super-resolution-based dual-path metric change detection network for remote sensing images with different resolutions},
  author={Li, Xi and Yan, Li and Zhang, Yi and Zeng, Huaien},
  journal={TGRS},
  year={2024}
}

@article{rsp,
  title={An empirical study of remote sensing pretraining},
  author={Wang, Di and Zhang, Jing and Du, Bo and Xia, Gui-Song and Tao, Dacheng},
  journal={TGRS},
  year={2022},
}

@article{saan,
  title={SAAN: Similarity-aware attention flow network for change detection with VHR remote sensing images},
  author={Guo, Haonan and Su, Xin and Wu, Chen and Du, Bo and Zhang, Liangpei},
  journal={TIP},
  year={2024}
}

@article{siamixformer,
  title={SiamixFormer: A fully-transformer Siamese network with temporal fusion for accurate building detection and change detection in bi-temporal remote sensing images},
  author={Mohammadian, Amir and Ghaderi, Foad},
  journal={International Journal of Remote Sensing},
  year={2023}
}

@article{transunetcd,
  title={TransUNetCD: A hybrid transformer network for change detection in optical remote-sensing images},
  author={Li, Qingyang and Zhong, Ruofei and Du, Xin and Du, Yu},
  journal={TGRS},
  year={2022}
}

@article{rdpnet,
  title={RDP-Net: Region Detail Preserving Network for Change Detection. arXiv 2022},
  author={Chen, H and Pu, F and Yang, R and Tang, R and Xu, X},
  year={2022},
  journal={arXiv}
}

@article{sdacd,
  title={An end-to-end supervised domain adaptation framework for cross-domain change detection},
  author={Liu, Jia and Xuan, Wenjie and Gan, Yuhang and Zhan, Yibing and Liu, Juhua and Du, Bo},
  journal={Pattern Recognition},
  year={2022}
}

@article{wnet,
  title={WNet: W-shaped hierarchical network for remote-sensing image change detection},
  author={Tang, Xu and Zhang, Tianxiang and Ma, Jingjing and Zhang, Xiangrong and Liu, Fang and Jiao, Licheng},
  journal={TGRS},
  year={2023}
}

@article{sgsln,
  title={Exchanging dual-encoder--decoder: A new strategy for change detection with semantic guidance and spatial localization},
  author={Zhao, Sijie and Zhang, Xueliang and Xiao, Pengfeng and He, Guangjun},
  journal={TGRS},
  year={2023}
}

@article{p2vcd,
  title={Transition is a process: Pair-to-video change detection networks for very high resolution remote sensing images},
  author={Lin, Manhui and Yang, Guangyi and Zhang, Hongyan},
  journal={TIP},
  year={2022}
}

@article{changeclip,
  title={ChangeCLIP: Remote sensing change detection with multimodal vision-language representation learning},
  author={Dong, Sijun and Wang, Libo and Du, Bo and Meng, Xiaoliang},
  journal={ISPRS},
  year={2024}
}

@article{nnunet,
  title={nnU-Net: a self-configuring method for deep learning-based biomedical image segmentation},
  author={Isensee, Fabian and Jaeger, Paul F and Kohl, Simon AA and Petersen, Jens and Maier-Hein, Klaus H},
  journal={Nature methods},
  year={2021}
}

@article{medmae,
  title={MedMAE: A Self-Supervised Backbone for Medical Imaging Tasks},
  author={Gupta, Anubhav and Osman, Islam and Shehata, Mohamed S and Braun, John W},
  journal={arXiv},
  year={2024}
}

@article{medsam,
  title={Segment Anything in Medical Images},
  author={Ma, Jun and He, Yuting and Li, Feifei and Han, Lin and You, Chenyu and Wang, Bo},
  journal={Nature Communications},
  year={2024}
}

@article{sammed2d,
  title={Sam on medical images: A comprehensive study on three prompt modes},
  author={Cheng, Dongjie and Qin, Ziyuan and Jiang, Zekun and Zhang, Shaoting and Lao, Qicheng and Li, Kang},
  journal={arXiv},
  year={2023}
}

@inproceedings{imissam,
  title={Interactive medical image segmentation: A benchmark dataset and baseline},
  author={Cheng, Junlong and Fu, Bin and Ye, Jin and Wang, Guoan and Li, Tianbin and Wang, Haoyu and Li, Ruoyu and Yao, He and Cheng, Junren and Li, JingWen and others},
  booktitle={CVPR},
  year={2025}
}

@article{cdmamba,
  title={CDMamba: Incorporating local clues into mamba for remote sensing image binary change detection},
  author={Zhang, Haotian and Chen, Keyan and Liu, Chenyang and Chen, Hao and Zou, Zhengxia and Shi, Zhenwei},
  journal={TGRS},
  year={2025}
}

@article{rsmamba,
  title={Rs-mamba for large remote sensing image dense prediction},
  author={Zhao, Sijie and Chen, Hao and Zhang, Xueliang and Xiao, Pengfeng and Bai, Lei and Ouyang, Wanli},
  journal={TGRS},
  year={2024}
}

@article{changemamba,
  title={ChangeMamba: Remote sensing change detection with spatiotemporal state space model},
  author={Chen, Hongruixuan and Song, Jian and Han, Chengxi and Xia, Junshi and Yokoya, Naoto},
  journal={TGRS},
  year={2024}
}

@article{whu,
  title={Fully convolutional networks for multisource building extraction from an open aerial and satellite imagery data set},
  author={Ji, Shunping and Wei, Shiqing and Lu, Meng},
  journal={TGRS},
  year={2018}
}

@article{LEVIR,
  title={A spatial-temporal attention-based method and a new dataset for remote sensing image change detection},
  author={Chen, Hao and Shi, Zhenwei},
  journal={Remote sensing},
  year={2020}
}

@article{SVCD,
  title={Change detection in remote sensing images using conditional adversarial networks},
  author={Lebedev, MA and Vizilter, Yu V and Vygolov, OV and Knyaz, Vladimir A and Rubis, A Yu},
  journal={The International Archives of the Photogrammetry, Remote Sensing and Spatial Information Sciences},
  year={2018}
}

@article{changen2,
  title={Changen2: Multi-temporal remote sensing generative change foundation model},
  author={Zheng, Zhuo and Ermon, Stefano and Kim, Dongjun and Zhang, Liangpei and Zhong, Yanfei},
  journal={TPAMI},
  year={2024}
}

@article{ye2024gmai,
  title={Gmai-mmbench: A comprehensive multimodal evaluation benchmark towards general medical ai},
  author={Ye, Jin and Wang, Guoan and Li, Yanjun and Deng, Zhongying and Li, Wei and Li, Tianbin and Duan, Haodong and Huang, Ziyan and Su, Yanzhou and Wang, Benyou and others},
  journal={NeurIPS},
  year={2024}
}

@article{openi,
  title={Preparing a collection of radiology examinations for distribution and retrieval},
  author={Demner-Fushman, Dina and Kohli, Marc D and Rosenman, Marc B and Shooshan, Sonya E and Rodriguez, Laritza and Antani, Sameer and Thoma, George R and McDonald, Clement J},
  journal={Journal of the American Medical Informatics Association},
  year={2016}
}

@inproceedings{huatuo,
  title={Towards injecting medical visual knowledge into multimodal llms at scale},
  author={Chen, Junying and Gui, Chi and Ouyang, Ruyi and Gao, Anningzhe and Chen, Shunian and Chen, Guiming and Wang, Xidong and Cai, Zhenyang and Ji, Ke and Wan, Xiang and others},
  booktitle={EMNLP},
  year={2024}
}

@inproceedings{lin2023pmc_oa,
  title={Pmc-clip: Contrastive language-image pre-training using biomedical documents},
  author={Lin, Weixiong and Zhao, Ziheng and Zhang, Xiaoman and Wu, Chaoyi and Zhang, Ya and Wang, Yanfeng and Xie, Weidi},
  booktitle={MICCAI},
  year={2023}
}

@article{zhang2023pmc_vqa,
  title={Pmc-vqa: Visual instruction tuning for medical visual question answering},
  author={Zhang, Xiaoman and Wu, Chaoyi and Zhao, Ziheng and Lin, Weixiong and Zhang, Ya and Wang, Yanfeng and Xie, Weidi},
  journal={arXiv},
  year={2023}
}

@inproceedings{hu2024omnimedvqa,
  title={Omnimedvqa: A new large-scale comprehensive evaluation benchmark for medical lvlm},
  author={Hu, Yutao and Li, Tianbin and Lu, Quanfeng and Shao, Wenqi and He, Junjun and Qiao, Yu and Luo, Ping},
  booktitle={CVPR},
  year={2024}
}

@article{ikezogwo2023quilt_1m,
  title={Quilt-1m: One million image-text pairs for histopathology},
  author={Ikezogwo, Wisdom and Seyfioglu, Saygin and Ghezloo, Fatemeh and Geva, Dylan and Sheikh Mohammed, Fatwir and Anand, Pavan Kumar and Krishna, Ranjay and Shapiro, Linda},
  journal={NeurIPS},
  year={2023}
}

@inproceedings{seyfioglu2024quilt_i,
  title={Quilt-llava: Visual instruction tuning by extracting localized narratives from open-source histopathology videos},
  author={Seyfioglu, Mehmet Saygin and Ikezogwo, Wisdom O and Ghezloo, Fatemeh and Krishna, Ranjay and Shapiro, Linda},
  booktitle={CVPR},
  year={2024}
}

@article{ji2022amos,
  title={Amos: A large-scale abdominal multi-organ benchmark for versatile medical image segmentation},
  author={Ji, Yuanfeng and Bai, Haotian and Ge, Chongjian and Yang, Jie and Zhu, Ye and Zhang, Ruimao and Li, Zhen and Zhanng, Lingyan and Ma, Wanling and Wan, Xiang and others},
  journal={NeurIPS},
  year={2022}
}

@article{brats,
  title={The 2024 brain tumor segmentation (brats) challenge: Glioma segmentation on post-treatment mri},
  author={de Verdier, Maria Correia and Saluja, Rachit and Gagnon, Louis and LaBella, Dominic and Baid, Ujjwall and Tahon, Nourel Hoda and Foltyn-Dumitru, Martha and Zhang, Jikai and Alafif, Maram and Baig, Saif and others},
  journal={arXiv},
  year={2024}
}

@misc{covid_quex,
	title={COVID-QU-Ex Dataset},
	DOI={10.34740/KAGGLE/DSV/3122958},
	publisher={Kaggle},
	author={Anas M. Tahir and Muhammad E. H. Chowdhury and Yazan Qiblawey and Amith Khandakar and Tawsifur Rahman and Serkan Kiranyaz and Uzair Khurshid and Nabil Ibtehaz and Sakib Mahmud and Maymouna Ezeddin},
	year={2022}
}

@inproceedings{s3d,
  title={Revisiting MAE pre-training for 3D medical image segmentation},
  author={Wald, Tassilo and Ulrich, Constantin and Lukyanenko, Stanislav and Goncharov, Andrei and Paderno, Alberto and Miller, Maximilian and Maerkisch, Leander and Jaeger, Paul and Maier-Hein, Klaus},
  booktitle={CVPR},
  year={2025}
}

@article{densevlm,
  title={Unbiased Region-Language Alignment for Open-Vocabulary Dense Prediction},
  author={Li, Yunheng and Li, Yuxuan and Zeng, Quansheng and Wang, Wenhai and Hou, Qibin and Cheng, Ming-Ming},
  journal={ICCV},
  year={2025}
}

@article{dontstop,
  title={Don't stop pretraining: Adapt language models to domains and tasks},
  author={Gururangan, Suchin and Marasovi{\'c}, Ana and Swayamdipta, Swabha and Lo, Kyle and Beltagy, Iz and Downey, Doug and Smith, Noah A},
  journal={arXiv},
  year={2020}
}

@inproceedings{resnet,
  title={Deep residual learning for image recognition},
  author={He, Kaiming and Zhang, Xiangyu and Ren, Shaoqing and Sun, Jian},
  booktitle={CVPR},
  year={2016}
}

@inproceedings{simmim,
  title={Simmim: A simple framework for masked image modeling},
  author={Xie, Zhenda and Zhang, Zheng and Cao, Yue and Lin, Yutong and Bao, Jianmin and Yao, Zhuliang and Dai, Qi and Hu, Han},
  booktitle={CVPR},
  year={2022}
}

@article{samrs,
  title={Samrs: Scaling-up remote sensing segmentation dataset with segment anything model},
  author={Wang, Di and Zhang, Jing and Du, Bo and Xu, Minqiang and Liu, Lin and Tao, Dacheng and Zhang, Liangpei},
  journal={NeurIPS},
  year={2023}
}

@article{zhang2022consecutive,
  title={Consecutive pre-training: A knowledge transfer learning strategy with relevant unlabeled data for remote sensing domain},
  author={Zhang, Tong and Gao, Peng and Dong, Hao and Zhuang, Yin and Wang, Guanqun and Zhang, Wei and Chen, He},
  journal={Remote Sensing},
  year={2022}
}

@inproceedings{flashattention,
  title={Flash{A}ttention: Fast and Memory-Efficient Exact Attention with {IO}-Awareness},
  author={Dao, Tri and Fu, Daniel Y. and Ermon, Stefano and Rudra, Atri and R{\'e}, Christopher},
  booktitle={NeurIPS},
  year={2022}
}

@article{ummae,
  title={Uniform masking: Enabling mae pre-training for pyramid-based vision transformers with locality},
  author={Li, Xiang and Wang, Wenhai and Yang, Lingfeng and Yang, Jian},
  journal={arXiv},
  year={2022}
}

@ARTICLE{11016924,
  author={Zhou, Jie and Liu, Yongxiang and Peng, Bowen and Liu, Li and Li, Xiang},
  journal={IEEE Transactions on Circuits and Systems for Video Technology}, 
  title={MaDiNet: Mamba Diffusion Network for SAR Target Detection}, 
  year={2025}}

@article{su2025rapid,
  title={Rapid Salient Object Detection With Difference Convolutional Neural Networks},
  author={Su, Zhuo and Liu, Li and M{\"u}ller, Matthias and Zhang, Jiehua and Wofk, Diana and Cheng, Ming-Ming and Pietik{\"a}inen, Matti},
  journal={TPAMI},
  year={2025},
}

@ARTICLE{li2024predicting,
  title = {Predicting gradient is better: Exploring self-supervised learning for SAR ATR with a joint-embedding predictive architecture},
  journal = {ISPRS},
  year = {2024},
  author = {Li, Weijie and Yang, Wei and Liu, Tianpeng and Hou, Yuenan and Li, Yuxuan and Liu, Zhen and Liu, Yongxiang and Liu, Li}}

@article{su2023lightweight,
  title={Lightweight pixel difference networks for efficient visual representation learning},
  author={Su, Zhuo and Zhang, Jiehua and Wang, Longguang and Zhang, Hua and Liu, Zhen and Pietik{\"a}inen, Matti and Liu, Li},
  journal={TPAMI},
  year={2023}
}

@article{hubel1962,
  title={Receptive fields, binocular interaction and functional architecture in the cat's visual cortex},
  author={Hubel, David H and Wiesel, Torsten N},
  journal={The Journal of physiology},
  year={1962}
}

@book{marr2010vision,
  title={Vision: A computational investigation into the human representation and processing of visual information},
  author={Marr, David},
  year={2010},
  publisher={MIT press}
}

@article{biederman1987,
  title={Recognition-by-components: a theory of human image understanding.},
  author={Biederman, Irving},
  journal={Psychological review},
  year={1987},
  publisher={American Psychological Association}
}

@article{kastner1998,
  title={Mechanisms of directed attention in the human extrastriate cortex as revealed by functional MRI},
  author={Kastner, Sabine and De Weerd, Peter and Desimone, Robert and Ungerleider, Leslie G},
  journal={science},
  year={1998}
}

@article{rao1999predictive,
  title={Predictive coding in the visual cortex: a functional interpretation of some extra-classical receptive-field effects},
  author={Rao, Rajesh PN and Ballard, Dana H},
  journal={Nature neuroscience},
  year={1999}
}

@article{muckli2015contextual,
  title={Contextual feedback to superficial layers of V1},
  author={Muckli, Lars and De Martino, Federico and Vizioli, Luca and Petro, Lucy S and Smith, Fraser W and Ugurbil, Kamil and Goebel, Rainer and Yacoub, Essa},
  journal={Current Biology},
  year={2015}
}

@inproceedings{aim,
  title={Multimodal autoregressive pre-training of large vision encoders},
  author={Fini, Enrico and Shukor, Mustafa and Li, Xiujun and Dufter, Philipp and Klein, Michal and Haldimann, David and Aitharaju, Sai and da Costa, Victor G Turrisi and B{\'e}thune, Louis and Gan, Zhe and others},
  booktitle={CVPR},
  year={2025}
}

@Article{mocov3,
  author  = {Xinlei Chen* and Saining Xie* and Kaiming He},
  title   = {An Empirical Study of Training Self-Supervised Vision Transformers},
  journal = {arXiv},
  year    = {2021},
}

@inproceedings{SynthDoG,
  title={Ocr-free document understanding transformer},
  author={Kim, Geewook and Hong, Teakgyu and Yim, Moonbin and Nam, JeongYeon and Park, Jinyoung and Yim, Jinyeong and Hwang, Wonseok and Yun, Sangdoo and Han, Dongyoon and Park, Seunghyun},
  booktitle={ECCV},
  year={2022}
}

@inproceedings{GeoQA,
  title={An augmented benchmark dataset for geometric question answering through dual parallel text encoding},
  author={Cao, Jie and Xiao, Jing},
  booktitle={International conference on computational linguistics}, 
  year={2022}
}

@inproceedings{sharegpt4v,
  title={Sharegpt4v: Improving large multi-modal models with better captions},
  author={Chen, Lin and Li, Jinsong and Dong, Xiaoyi and Zhang, Pan and He, Conghui and Wang, Jiaqi and Zhao, Feng and Lin, Dahua},
  booktitle={ECCV},
  year={2024}
}

@inproceedings{dvqa,
  title={Dvqa: Understanding data visualizations via question answering},
  author={Kafle, Kushal and Price, Brian and Cohen, Scott and Kanan, Christopher},
  booktitle={CVPR},
  year={2018}
}

@article{chartqa,
  title={Chartqa: A benchmark for question answering about charts with visual and logical reasoning},
  author={Masry, Ahmed and Long, Do Xuan and Tan, Jia Qing and Joty, Shafiq and Hoque, Enamul},
  journal={ACL},
  year={2022}
}

@inproceedings{ai2d,
  title={A diagram is worth a dozen images},
  author={Kembhavi, Aniruddha and Salvato, Mike and Kolve, Eric and Seo, Minjoon and Hajishirzi, Hannaneh and Farhadi, Ali},
  booktitle={ECCV},
  pages={235--251},
  year={2016},
  organization={Springer}
}

@inproceedings{docvqa,
  title={Docvqa: A dataset for vqa on document images},
  author={Mathew, Minesh and Karatzas, Dimosthenis and Jawahar, CV},
  booktitle={Winter conference on applications of computer vision},
  pages={2200--2209},
  year={2021}
}

@article{comp,
  title={Comp: Continual multimodal pre-training for vision foundation models},
  author={Chen, Yitong and Meng, Lingchen and Peng, Wujian and Wu, Zuxuan and Jiang, Yu-Gang},
  journal={arXiv preprint arXiv:2503.18931},
  year={2025}
}

@article{llavasiglip,
  title={Llava-onevision: Easy visual task transfer},
  author={Li, Bo and Zhang, Yuanhan and Guo, Dong and Zhang, Renrui and Li, Feng and Zhang, Hao and Zhang, Kaichen and Zhang, Peiyuan and Li, Yanwei and Liu, Ziwei and others},
  journal={arXiv preprint arXiv:2408.03326},
  year={2024}
}

@inproceedings{siglip,
  title={Sigmoid loss for language image pre-training},
  author={Zhai, Xiaohua and Mustafa, Basil and Kolesnikov, Alexander and Beyer, Lucas},
  booktitle={ICCV},
  pages={11975--11986},
  year={2023}
}

@inproceedings{aimv2,
  title={Multimodal autoregressive pre-training of large vision encoders},
  author={Fini, Enrico and Shukor, Mustafa and Li, Xiujun and Dufter, Philipp and Klein, Michal and Haldimann, David and Aitharaju, Sai and da Costa, Victor G Turrisi and B{\'e}thune, Louis and Gan, Zhe and others},
  booktitle={CVPR},
  year={2025}
}

@inproceedings{deit3,
  title={Deit iii: Revenge of the vit},
  author={Touvron, Hugo and Cord, Matthieu and J{\'e}gou, Herv{\'e}},
  booktitle={European conference on computer vision},
  pages={516--533},
  year={2022},
  organization={Springer}
}

@article{4m,
  title={4m: Massively multimodal masked modeling},
  author={Mizrahi, David and Bachmann, Roman and Kar, Oguzhan and Yeo, Teresa and Gao, Mingfei and Dehghan, Afshin and Zamir, Amir},
  journal={Advances in Neural Information Processing Systems},
  volume={36},
  pages={58363--58408},
  year={2023}
}

@inproceedings{coco,
  title={Microsoft coco: Common objects in context},
  author={Lin, Tsung-Yi and Maire, Michael and Belongie, Serge and Hays, James and Perona, Pietro and Ramanan, Deva and Doll{\'a}r, Piotr and Zitnick, C Lawrence},
  booktitle={ECCV},
  year={2014}
}

@inproceedings{ade20k,
  title={Scene parsing through ade20k dataset},
  author={Zhou, Bolei and Zhao, Hang and Puig, Xavier and Fidler, Sanja and Barriuso, Adela and Torralba, Antonio},
  booktitle={CVPR},
  year={2017}
}

@article{darcet2023vision,
  title={Vision transformers need registers},
  author={Darcet, Timoth{\'e}e and Oquab, Maxime and Mairal, Julien and Bojanowski, Piotr},
  journal={arXiv preprint arXiv:2309.16588},
  year={2023}
}

@article{jiang2025vision,
  title={Vision Transformers Don't Need Trained Registers},
  author={Jiang, Nick and Dravid, Amil and Efros, Alexei and Gandelsman, Yossi},
  journal={arXiv preprint arXiv:2506.08010},
  year={2025}
}
}


\end{document}